\pdfoutput=1
\documentclass[11pt]{article}

\usepackage[final]{ACL2023}

\usepackage{times}
\usepackage{latexsym}

\usepackage[T1]{fontenc}
\usepackage[utf8]{inputenc}

\usepackage{microtype}

\usepackage{inconsolata}

\newcommand{\quotes}[1]{``#1''}
\usepackage{amsmath}
\usepackage{tikz}
 
\usepackage[export]{adjustbox}
\usepackage{booktabs}
\usepackage{multirow}
\usepackage{hhline}

\usepackage{hyperref}
\usepackage{arydshln}

\usepackage{pifont}% http://ctan.org/pkg/pifont
\usepackage{colortbl}

\usepackage{enumitem}
\setlist[itemize]{noitemsep, topsep=3pt, partopsep=0pt, parsep=0pt}

\title{CodeNER: Code Prompting for Named Entity Recognition}

\author{Sungwoo Han$^{1\dagger}$, Hyeyeon Kim$^{1\dagger}$, $^*$Jingun Kwon$^1$, \\ \textbf{Hidetaka Kamigaito}$^2$, \textbf{and Manabu Okumura}$^3$ \\
 $^1$Chungnam National University, $^2$Nara Institute of Science and Technology (NAIST) \\
 $^3$Institute of Science Tokyo \\
 {\tt \{77sungwhan,hyk22\}@o.cnu.ac.kr} \\
 {\tt jingun.kwon@cnu.ac.kr} \\
 {\tt kamigaito.h@is.naist.jp} \\
 {\tt oku@pi.titech.ac.jp}
 \\}
\begin{document}
\maketitle
\begin{abstract}
Recent studies have explored various approaches for treating candidate named entity spans as both source and target sequences in named entity recognition (NER) by leveraging large language models (LLMs). Although previous approaches have successfully generated candidate named entity spans with suitable labels, they rely solely on input context information when using LLMs, particularly, ChatGPT. However, NER inherently requires capturing detailed labeling requirements with input context information. To address this issue, we propose a novel method that leverages code-based prompting to improve the capabilities of LLMs in understanding and performing NER. By embedding code within prompts, we provide detailed BIO schema instructions for labeling, thereby exploiting the ability of LLMs to comprehend long-range scopes in programming languages. Experimental results demonstrate that the proposed code-based prompting method outperforms conventional text-based prompting on ten benchmarks across English, Arabic, Finnish, Danish, and German datasets, indicating the effectiveness of explicitly structuring NER instructions. We also verify that combining the proposed code-based prompting method with the chain-of-thought prompting further improves performance.\footnote{Our code and dataset are available at \url{https://github.com/XXXX}.}  
\end{abstract}

\section{Introduction}
Named entity recognition (NER) is a task of detecting and classifying named entities in a given text, such as \textit{person}, \textit{location}, and \textit{miscellaneous} entities. Traditionally, NER has been treated as a sequential labeling problem, e.g., with the BIO output schema, where a learned model classifies each token in a sequence from a predefined label set~\cite{carreras-etal-2003-learning,lample-etal-2016-neural,ma-hovy-2016-end,wan-etal-2022-nested}. These schemata indicate that each token in the input should be aligned with the BIO tag by capturing entity boundaries.   
Incorporating conditional random field (CRF) networks, which explicitly capture broader context information of input, has led to leading performance~\cite{10.5555/645530.655813,finkel-etal-2005-incorporating,liu-etal-2011-recognizing,lample-etal-2016-neural,jie-lu-2019-dependency,kruengkrai-etal-2020-improving,xu-etal-2021-better}.

Recently, large language models (LLMs) have achieved great success in various natural language processing (NLP) tasks by leveraging in-context learning with prompts and exploiting their zero-shot task-solving abilities without the need to update model parameters~\cite{neuips2022,weifinetuned,51115}. There are two main types of LLMs: closed and open. While closed models, such as ChatGPT, which is based on GPT-4,\footnote{\url{https://chat.openai.com/}} have shown strong reasoning abilities, open models, such as Llama-3
% \footnote{\url{https://huggingface.co/meta-llama/Meta-Llama-3-8B-instruct}} 
and Phi-3,
% \footnote{\url{https://huggingface.co/microsoft/Phi-3-mini-128k-instruct}} 
offer flexibility to access and modify their architectures~\cite{llama3modelcard,abdin2024phi3technicalreporthighly}. 

Motivated by these recent advancements, the zero-shot reasoning abilities of LLMs, specifically GPT, in NER have been explored~\cite{xie-etal-2023-empirical,10.1109/JCDL57899.2023.00034,xie-etal-2024-self,10.1007/978-3-031-72437-4_22}. Despite the success of using LLMs, however, previous work has relied on two-stage approaches, such as majority voting or constructing self-annotated datasets, for few-shot techniques with text-based prompting. Text-based prompting faces challenges in NER, primarily due to the inherent gap between the text-in-text-out schema of LLMs and the text-in-span-out nature of the task~\cite{10.5555/AAI28114605}. 
% challenge and need to identify entity boundaries
This mismatch makes it difficult to handle the sequential aspect of the BIO output schema, which is integral to NER, in LLMs and can result in poor performance in identifying entity boundaries~\cite{ding-etal-2024-rethinking}. Although various prompting methods, such as chain-of-thought (CoT)~\cite{han2023informationextractionsolvedchatgpt,xie-etal-2023-empirical}, have been introduced to improve reasoning~\cite{cot}, challenges still remain in bridging this gap.

In this paper, we propose CodeNER, a code-based prompting method designed to improve the capabilities of LLMs in understanding and performing NER. While previous supervised frameworks learn an explicit mapping from each token to its label, this mapping is not captured in zero-shot and few-shot settings, specifically in-context learning.   
By embedding detailed BIO schema labeling instructions within complete, structured code prompts, CodeNER better identifies entity boundaries by exploiting the LLMs' inherent ability to comprehend programming language structures. Unlike previous text-based or partial code-based approaches, our method explicitly guides the sequential processing essential for accurate NER, thereby bridging the gap between the text-in-text-out schema of LLMs and the text-in-span-out nature of the task. This explicit sequential guidance is particularly crucial in zero-shot and few-shot scenarios, where explicit supervision is minimal. Completely structured code prompts with BIO out schema inherently capture necessary sequential context, addressing the limitations of zero-shot and few-shot methods, which typically lack explicit sequential modeling due to the absence of direct token-label supervision.

We evaluate the performance of CodeNER compared with the text-based prompting methods for both closed models of ChatGPT (GPT-4 and GPT-4 Turbo) and open models of Llama-3-8B and Phi-3-mini-128k-instruct. 
Experimental results obtained on ten benchmarks across English, Arabic, Finnish, Danish, and German NER datasets demonstrated that the proposed CodeNER outperforms the text-based prompting methods. We perform in-depth analysis to evaluate the effectiveness of CodeNER in terms of overcoming the limitations of the text-based prompting methods, that rely solely on context information of input in zero-shot settings. We also show that the performance of the proposed method can be further improved by incorporating CoT into CodeNER.

\section{Related Work}
\textbf{Named Entity Recognition.}
NER is the task of identifying and annotating named entities in a given text from a predefined label set. Traditionally, NER has been approached as a sequential labeling task~\cite{carreras-etal-2003-learning,lample-etal-2016-neural}. Early efforts employed LSTM networks to represent token embeddings and subsequently integrated conditional random field (CRF) layers with CNNs to consider broader context information~\cite{collobert}. In addition, hybrid models that combine LSTM and CNNs have been investigated to leverage character-level information~\cite{chiu-nichols-2016-named,ma-hovy-2016-end}. Another approach formulates the NER task as span prediction, where learned models simultaneously detect span boundaries and annotate named entities~\cite{stratos-2017-entity,li-etal-2020-unified,ouchi-etal-2020-instance,fu-etal-2021-spanner}. Recent work has extended the span-based approaches to better capture nested entities, which can appear when one named entity is contained within another~\cite{shen-etal-2021-locate,tan2021sequencetoset,ding-etal-2022-ask}. Instead of formulating NER as a sequential labeling task, generative methods have been introduced, that utilize a Seq2Seq framework by considering candidate named entity spans as both source and target sequences~\cite{cui-etal-2021-template,Fei_Ji_Li_Liu_Ren_Li_2021,yan-etal-2021-unified-generative}. While supervised methods have achieved saturated performance on the standard NER benchmarks, they require extensive labeled datasets~\cite{chen-etal-2021-data,zhou-etal-2022-melm} and domain adaptation methods~\cite{zhang-etal-2021-pdaln,yang-etal-2022-factmix,xu-etal-2023-improving}.

\noindent \textbf{In-context Learning.}
LLMs have achieved significant success in various NLP %natural language processing 
tasks in zero-shot settings by leveraging specifically formatted instruction-based inputs~\cite{Radford2019LanguageMA,NEURIPS2020_1457c0d6,chowdhery2022palm}.
Different from the fine-tuning setting, which requires additional parameter updates for learning to solve tasks, the zero-shot setting leverages in-context learning to perform tasks without any additional adjustments of the parameters of a pre-trained model. Owing to its flexibility, numerous efforts have been made to investigate better instructions for enhancing reasoning abilities, including few-shot, directional stimulus, and CoT learning~\cite{liu-etal-2022-generated,zhu2023multilingual,cot,li2023guiding}.
Recently, the use of LLMs in NER has gained attention due to their remarkable reasoning abilities. \newcite{wang2023gptnernamedentityrecognition} considered a retrieval-augmented generation approach by leveraging entity-level information using a fine-tuned NER model to search exemplars. \newcite{xie-etal-2023-empirical} explored the reasoning abilities of ChatGPT by considering few-shot techniques based on a two-stage majority voting. \newcite{xie-etal-2024-self} demonstrated that self-annotating an unlabeled dataset using ChatGPT further improves performance with the two-stage majority voting. Additionally, instruction-based fine-tuning with LLMs and evaluating unseen labels on new datasets using different prompts have also been studied~\cite{sainz2024gollie,ding-etal-2024-rethinking,zamai2024lessinstructmoreenriching}.

\begin{figure*}[ht!]
  \centering
  \includegraphics[width=0.73\linewidth]{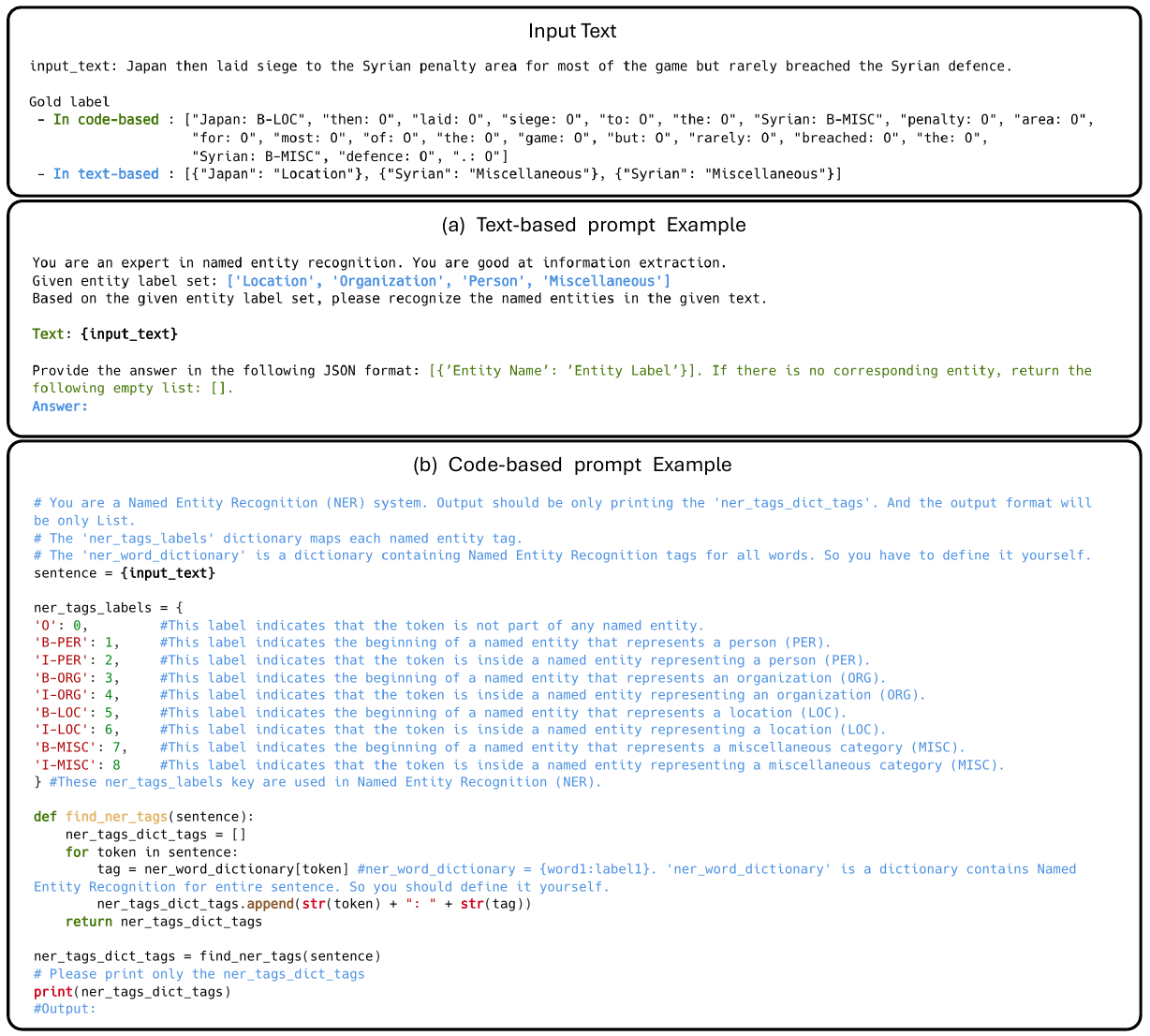}
  \caption{Examples of text- and code-based prompts for zero-shot NER with ChatGPT. (a) and (b) indicate the text-based and code-based prompts, respectively.}
  \label{fig:codeBasedPrompt}
\end{figure*}

In this study, instead of leveraging a text-based approach, we focus on leveraging a programming language-style for NER to exploit the structured nature inherent in code. Previous studies have primarily focused on broader context information of input and have not thoroughly explored the sequential aspect of BIO output schema into NER prompting. The proposed CodeNER bridges this gap by integrating the BIO output schema and broader context of input in LLMs to improve NER performance.

\section{CodeNER}
Unlike text-based prompting methods, which rely solely on natural language instructions for NER, our CodeNER is a code-based prompting method. %It is specifically designed to bridge the gap between the text-in-text-out schema of LLMs and text-in-span-out nature} in LLMs within a single prompt. 
Figures~\ref{fig:codeBasedPrompt} (a) and (b) show text-based prompts and code-based prompts in Python, respectively.
CodeNER explicitly encourages models to consider the BIO output schema and broader context information of input via prompts written in a programming language-style. 
This approach is critical for improving performance in NER tasks since it enables the model to better capture entity boundaries~\cite{ding-etal-2024-rethinking} than existing methods using LLMs that consider only context information~\cite{liu-etal-2011-recognizing,kruengkrai-etal-2020-improving,xu-etal-2021-better}.
Python code-based prompts provide a clear and human-readable structure while overcoming several challenges, such as potential misinterpretation and variability in natural languages~\cite{wang-etal-2023-code4struct,LiSTYWHQ23,sainz2024gollieannotationguidelinesimprove, zeng-etal-2025-codetaxo}. 

% Figures~\ref{fig:codeBasedPrompt} (a) and (b) show text-based prompts and code-based prompts in Python, respectively. The text-based prompt relies solely on natural language instructions for NER, whereas the code-based prompt leverages the structured nature of programming languages to explicitly guide the model by bridging the gap between traditional \textcolor{red}{token} prediction and the generative properties of LLMs. 

In CodeNER, the variable \quotes{sentence} contains a list of sentences tokenized for processing, and \quotes{ner\_tag\_labels} is a dictionary that maps each NER label to a numeric identifier with comments that explain each label. The prompt includes a function that performs NER, followed by a print statement that outputs the annotated named entities. The function \quotes{find\_ner\_tags} uses a for-loop to guide the LLM to process the BIO output schema and input sentence information. \footnote{We performed preliminary tests and found that the function is indeed essential for improving performance and facilitating accurate parsing.} In addition, \quotes{ner\_word\_dictionary,} which is not predefined, is included with explanatory comments to prompt the model to define it dynamically for NER. The classified results are appended to the list \quotes{ner\_tags\_dict\_tags,} and the final output is generated.

By embedding the prompt within a code structure, CodeNER not only specifies the BIO output schema but also incorporates input sentence information, providing a comprehensive framework for the model to perform NER.

\section{Experiments}
\subsection{Experimental Settings}

\begin{table}[t!]
\renewcommand{\arraystretch}{1}
% \begin{adjustbox}{width=0.88\columnwidth, center}
\centering
\Huge
\resizebox{0.5\textwidth}{!}{
\begin{tabular}{cccccccc}
\rowcolor{gray!10}
\toprule

\textbf{Language} &\textbf{Dataset} & \textbf{\#Sent.} & \textbf{\#Label} & \textbf{Language} &\textbf{Dataset} & \textbf{\#Sent.} & \textbf{\#Label}\\
\midrule
\multirow{5}{*}{\centering English} & CoNLL03 & 3,453 & 5 & \multirow{2}{*}{\centering Arabic}  & Arabic A & 267 & 5 \\
        
        & FIN       & 305 & 5 &  & Arabic B & 254 & 5 \\
       & MIT Movie Trivia & 1,953 & 13  & Finnish & finnish A & 986 & 5 \\ 
        & MIT restaurant   & 1,521 & 9 & Danish  & DaNE & 565 & 5  \\
        & WNUT 17          & 1,287 & 7   & German  & SwissNER & 200 & 5 \\

\bottomrule
\end{tabular}}
% \end{adjustbox}

\caption{Statistics of the datasets. \#Sent. and \#label indicate the dataset size and the number of named entities, respectively.}
\label{tab:datasetInfo}
\end{table}

\noindent \textbf{Datasets.}
We used ten benchmark datasets. Table~\ref{tab:datasetInfo} provides the statistics of the NER test datasets. 
For the English datasets, we used CoNLL03~\cite{tjong-kim-sang-de-meulder-2003-introduction}, FIN~\cite{salinas-alvarado-etal-2015-domain}, WNUT-17~\cite{derczynski-etal-2017-results}, the MIT movie review semantic corpus, and the MIT restaurant review.\footnote{The MIT NER datasets are publicly available  at \url{https://groups.csail.mit.edu/sls/downloads/}.} For the Arabic datasets, we used ArabicDatasets A and B~\cite{bari19}. For the Finnish dataset, we used FinnishDataset A~\cite{Ruokolainen_2019}. The German and Danish datasets are SwissNER~\cite{vamvas-etal-2023-swissbert} and DaNE~\cite{hvingelby-etal-2020-dane}, respectively. 

We %constructed new datasets by 
transformed the traditional NER datasets into a programming language-style format. Note that, when considering languages other than English, we used Google Translate to translate the \quotes{ner\_tags\_labels} descriptions in both CodeNER and Vanilla prompts.\footnote{\url{https://translate.google.com/}} 
Different from previous work that sampled only a small portion of datasets~\cite{xie-etal-2023-empirical,xie-etal-2024-self}, we used the entire datasets to ensure a comprehensive evaluation.

\noindent \textbf{Implementation Details.}
For closed models, we employed the recent ChatGPT (GPT-4 and GPT-4 Turbo) models with their official APIs. For open models, we employed open-source Llama-3-8B and Phi-3-mini-128k-instruct as the backbone models for our experiments. Following previous work, we used F$_1$ scores across all datasets based on the model-generated outputs~\cite{cui-etal-2021-template,xie-etal-2023-empirical} by considering the traditional evaluation method of BIO tagging~\cite{hammerton-2003-named,ma-hovy-2016-end,chen-etal-2020-local}.  All experiments for the open models were conducted using an NVIDIA RTX A6000 GPU. For the decoding step, we set the $\rm do_{sample}$ option to \texttt{False}.\footnote{All hyperparameter settings are described in Appendix~\ref{appen:hyp}.}

\noindent \textbf{Baseline.} 
We compared the proposed CodeNER to vanilla text-based prompting NER, shown in Figure~\ref{fig:codeBasedPrompt} (a)~\cite{xie-etal-2023-empirical} in zero-shot settings.\footnote{ We converted model-generated outputs to the BIO format for evaluation.}

\subsection{Main Results}
\begin{table}[t!]
\renewcommand{\arraystretch}{1}
% \begin{adjustbox}{width=1\columnwidth,center}
\centering
\Huge
\resizebox{0.5\textwidth}{!}{
\begin{tabular}{cccccccc}
\toprule
\rowcolor{gray!10}
& & \multicolumn{3}{c}{\textbf{GPT-4}} & \multicolumn{3}{c}{\textbf{GPT-4 turbo}} \\
\rowcolor{gray!10}
\cline{3-5} \cline{6-8}
% \cmidrule(lr){3-5} \rowcolor{gray!10}\cmidrule(lr){6-8}
\rowcolor{gray!10} 
\multirow{-2}{*}{\centering \textbf{Language}} & \multirow{-2}{*}{\centering \textbf{Dataset}} & \textbf{Vanilla} & \textbf{CodeNER} & \textbf{$\Delta$} & \textbf{Vanilla} & \textbf{CodeNER} & \textbf{$\Delta$}\\

\midrule
%&&&{\textbf{R-1}}&{\textbf{R-2}}&\textbf{{R-L}}&&&\\ 

\multirow{5}{*}{\centering English} & CoNLL03 & 73.24 & 69.48 & {-3.76} & 74.54 & 68.31 & {-6.23} \\
            &FIN                & 22.83 & 41.13 & \textbf{18.30}& 11.58 & 27.45 & \textbf{15.87} \\
            &MIT restaurant     & 40.47 & 49.95 & \textbf{9.48}& 36.34 & 46.05 & \textbf{9.71} \\
            &MIT movie    & 52.21 & 48.38 &{-3.83}& 50.78 & 46.80 & {-3.98} \\
            &WNUT-17            & 51.04 & 46.00 & {-5.04} & 42.80 & 52.09 & \textbf{9.29}\\
\midrule
\multirow{2}{*}{\centering Arabic} &Arabic A     & 38.31 & 44.52 & \textbf{6.21} & 39.11 & 41.58 & \textbf{2.47} \\
            &Arabic B     & 43.04 & 45.77 & \textbf{2.73} & 44.29 & 46.57 & \textbf{2.28} \\
\midrule
Finnish     &Finnish A    & 54.26 & 59.85 & \textbf{5.59} & 49.32 & 62.98 & \textbf{13.66} \\
\midrule
German      &SwissNER           & 58.18 & 61.74 & \textbf{3.56} & 66.71 & 57.22 & {-9.49} \\
\midrule
Danish      &DaNE               & 66.46 & 57.28 & {-9.18} & 65.53 & 62.61 & {-2.92} \\
\midrule
            &     Average            & 50.00 & 52.41 & \textbf{2.41}  & 48.10  & 51.17 & \textbf{3.07} \\
\bottomrule

  \end{tabular}}

% \end{adjustbox}

\caption{Experimental results with macro-F1 scores on ten benchmarks. $\Delta$ indicates the performance difference between CodeNER and Vanilla.
}
\label{tab:MaintotalResult}
\end{table}

\begin{table*}[t!]
% \begin{adjustbox}{width=1.65\columnwidth,center}
\renewcommand{\arraystretch}{0.8}
\centering
\Huge
\resizebox{0.85\textwidth}{!}{
\begin{tabular}{cccccccccccccc}
\rowcolor{gray!10}
\toprule
& & & \textbf{PER} & & & \textbf{LOC} & & & \textbf{ORG} & & & \textbf{MISC} & \\
% \cmidrule(lr){3-5} \cmidrule(lr){6-8}
% \cmidrule(lr){9-11} \cmidrule(lr){12-14} 

\cline{3-14}
\rowcolor{gray!10}
\multirow{-2}{*}{\centering \textbf{Dataset}} & \multirow{-2}{*}{\centering \textbf{Model}}  & \textbf{Vanilla} & \textbf{CodeNER} & \textbf{$\Delta$} & \textbf{Vanilla} & \textbf{CodeNER} & \textbf{$\Delta$} & \textbf{Vanilla} & \textbf{CodeNER} & \textbf{$\Delta$} & \textbf{Vanilla} & \textbf{CodeNER} & \textbf{$\Delta$} \\ 
\midrule

\multirow{2}{*}{\centering CoNLL03} & GPT-4       & 92.06 & 88.38 & {-3.68} & 79.45 & 75.77 & {-3.68} & 66.15 & 58.53 & {-7.62} & 31.88 & 36.87 & \textbf{4.99} \\
                & GPT-4 turbo & 90.14 & 86.16 & {-3.98} & 83.19 & 72.19 & {-11.00} & 63.96 & 55.20 & {-8.76} & 43.07 & 48.99 & \textbf{5.92} \\
\midrule
\multirow{2}{*}{\centering FIN} & GPT-4       & 26.94 & 51.71 & \textbf{24.77} & 15.62 & 38.10 & \textbf{22.48} & 14.72 & 7.48 & {-7.24} &  1.06 & 0.78 & {-0.28}\\
                & GPT-4 turbo &  5.29 & 31.76 & \textbf{26.47} & 30.77 & 27.72 & {-3.05} & 23.93 & 13.98 & {-9.95} &  0.00 & 0.79 & \textbf{0.79} \\
\midrule
\multirow{2}{*}{\centering Arabic A}  & GPT-4       & 63.20 & 71.10 & \textbf{7.90} & 52.99 & 55.64 & \textbf{2.65} & 21.38 & 15.31 & {-6.07} & 16.91 & 26.53 & \textbf{9.62} \\
                & GPT-4 turbo & 54.08 & 68.33 & \textbf{14.25} & 59.06 & 57.30 & {-1.76} & 27.37 & 18.39 & {-8.98} & 18.90 & 19.42 & \textbf{0.52} \\
\midrule
\multirow{2}{*}{\centering Arabic B}  & GPT-4       & 63.97 & 67.94 & \textbf{3.97} & 57.43 & 59.73 & \textbf{2.30} & 21.62 & 21.05 & {-0.57} & 23.66 & 26.42 & \textbf{2.76} \\
                & GPT-4 turbo & 62.70 & 68.47 & \textbf{5.77} & 61.68 & 64.18 & \textbf{2.50} & 27.27 & 26.75 & {-0.52} & 24.07 & 24.26 & \textbf{0.19} \\
\midrule
\multirow{2}{*}{\centering Finnish A} & GPT-4       & 73.57 & 62.48 & {-11.09} & 63.81 & 81.83 & \textbf{18.02} & 60.20 & 66.87 & \textbf{6.67} & 12.83 & 14.73 & \textbf{1.90} \\
                & GPT-4 turbo & 65.75 & 82.32 & \textbf{16.57} & 62.88 & 79.84 & \textbf{16.96} & 53.78 & 68.06 & \textbf{14.28} &  9.02 & 14.92 & \textbf{5.90} \\
\midrule
\multirow{2}{*}{\centering SwissNER}        & GPT-4       & 65.00 & 72.66 & \textbf{7.66} & 75.26 & 72.30 & {-2.96} & 60.75 & 63.23 & \textbf{2.48} & 20.00 & 32.16 & \textbf{12.16}\\
                & GPT-4 turbo & 83.00 & 74.07 & {-8.93} & 80.00 & 71.39 & {-8.61} & 73.96 & 52.36 & {-21.60} & 17.93 & 27.18 & \textbf{9.25} \\
\midrule
\multirow{2}{*}{\centering DaNE}            & GPT-4       & 85.13 & 63.67 & {-21.46} & 76.15 & 66.12 & {-10.03} & 70.49 & 60.70 & {-9.79} & 25.66 & 36.20 & \textbf{10.54} \\
                & GPT-4 turbo & 83.91 & 79.77 & {-4.14} & 85.56 & 69.23 & {-16.33} & 67.91 & 63.19 & {-4.72} & 19.12 & 31.06 & \textbf{11.94} \\
\midrule
\multirow{2}{*}{\centering Average}                & GPT-4       & 67.12 & 68.28 & \textbf{1.16} & 60.10 & 64.21 & \textbf{4.11} & 45.04 & 41.88 & {-3.16} & 18.86 & 24.81 & \textbf{5.95} \\
                & GPT-4 turbo & 63.55 & 70.13 & \textbf{6.58} & 66.16 & 63.12 & {-3.04} & 48.31 & 42.56 & {-5.75} & 18.87 & 23.80 & \textbf{4.93} \\

\bottomrule
\end{tabular}}

% \end{adjustbox}

\caption{Experimental results with F1-scores for each label on \texttt{CoNLL03}, \texttt{FIN}, \texttt{ArabicDatasetA}, \texttt{ArabicDatasetB}, \texttt{finnishDatasetA}, \texttt{DaNE}, and \texttt{SwissNER}, where the label set includes  {PER}, {LOC}, {ORG}, and {MISC}.}

\label{tab:FiveLabelResult}
\end{table*}

Table~\ref{tab:MaintotalResult} shows the zero-shot results of CodeNER and Vanilla using GPT-4 and GPT-4 Turbo. For GPT-4, CodeNER outperformed the Vanilla method on most datasets, particularly with a significant improvement on \texttt{FIN}, where CodeNER achieved an 18.3 point improvement in the $F_1$ score. Similarly, a significant improvement was obtained on \texttt{MIT restaurant}, with an increase of 9.48 points. However, CodeNER underperformed GPT-4 on \texttt{CoNLL03} and \texttt{WNUT-17}. On average, CodeNER improved the $F_1$ score by 2.41 points, representing a 4.82\% improvement across all datasets. For GPT-4 turbo, similar performance gains were obtained by CodeNER, with an average improvement of 3.07 points in the $F_1$ score, representing a 6.37\% improvement across all datasets. 

Table~\ref{tab:FiveLabelResult} shows a detailed comparison between CodeNER and Vanilla across four NER labels: PER (Person), LOC (Location), ORG (Organization), and MISC (Miscellaneous) on \texttt{CoNLL03}, \texttt{FIN}, \texttt{Arabic A}, \texttt{Arabic B}, \texttt{Finnish A}, \texttt{DaNE}, and \texttt{SwissNER}. On average, the proposed CodeNER generally outperformed Vanilla, except for the ORG label. In particular, we observed that CodeNER consistently outperforms Vanilla for the MISC label. Although there exist labels for which CodeNER underforms ChatGPT on certain datasets, specifically \texttt{CoNLL03} and \texttt{SwissNER}, the overall performance demonstrated that CodeNER, which can incorporate the BIO output schema together, is more robust for NER in zero-shot settings. Table~\ref{tab:few_shot_settings} shows the additional results for one-shot and few-shot settings.

\begin{figure}[t!]
  \centering
  \includegraphics[width=1\linewidth]{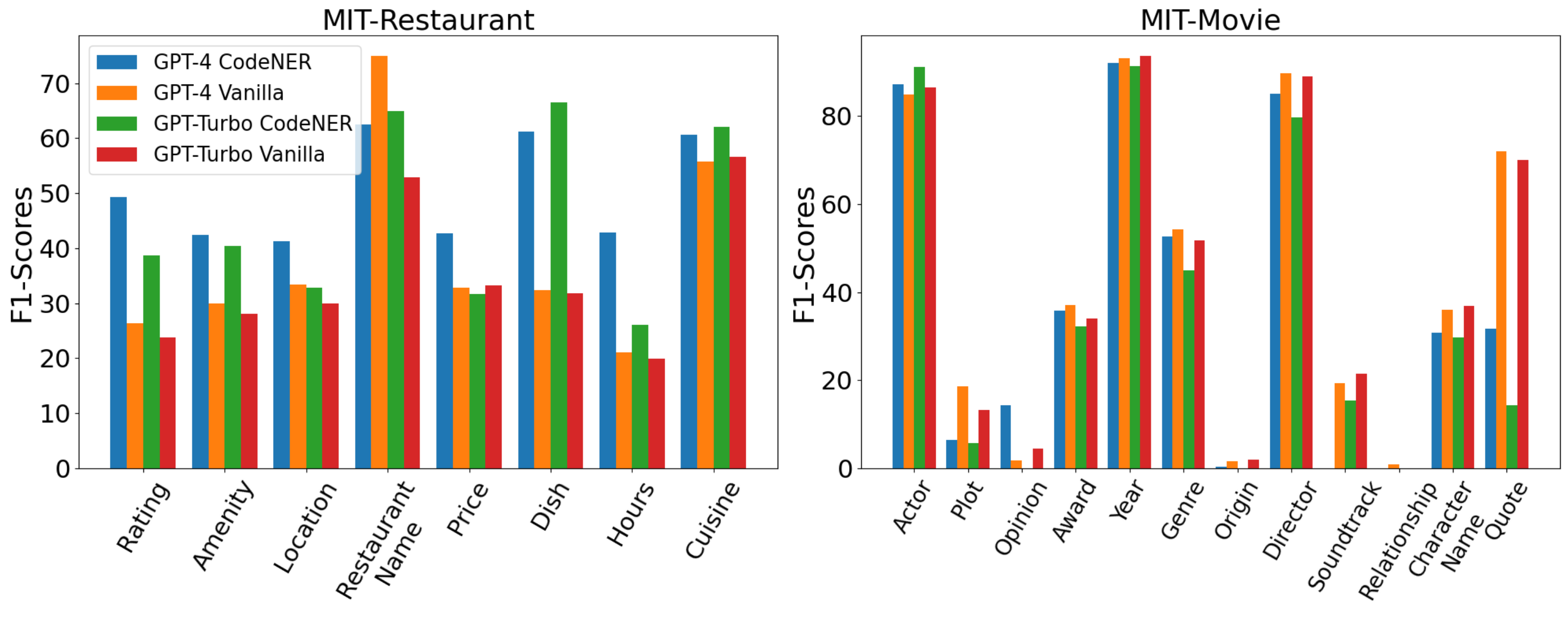}
  \caption{Experimental results for each label on \texttt{MIT-Restaurant} and \texttt{MIT-Movie}.}
  \label{fig:MIT-scores}
\end{figure}

Figure~\ref{fig:MIT-scores} shows a detailed comparison for each label between CodeNER and Vanilla on \texttt{MIT-restaurant} and \texttt{MIT-movie}. On \texttt{MIT-restaurant}, CodeNER consistently outperformed Vanilla across most of the labels. However, on \texttt{MIT-movie}, Vanilla outperformed CodeNER for the \quotes{Plot} and \quotes{Quote} labels even though CodeNER maintained competitive scores.

\begin{table*}[t!]
% \begin{adjustbox}{width=1.85\columnwidth,center}
\renewcommand{\arraystretch}{0.8}
\centering
\resizebox{0.85\textwidth}{!}{

\Huge
\begin{tabular}{cccccccccccccccccccc}
\toprule
\rowcolor{gray!10}
\textbf{Model} & \textbf{Dataset} & \textbf{Template} & {\textbf{$F_1$}} & {\textbf{PER}} & {\textbf{LOC}} &{\textbf{ORG}} & {\textbf{MISC}}  & \textbf{Model} & {\textbf{$F_1$}} & {\textbf{PER}} & {\textbf{LOC}} &{\textbf{ORG}} & {\textbf{MISC}} & \textbf{Model} & {\textbf{$F_1$}} & {\textbf{PER}} & {\textbf{LOC}} &{\textbf{ORG}} & {\textbf{MISC}}\\
\midrule
\multirow{41.5}{*}{\centering Phi-3} & \multirow{5}{*}{\centering FIN} & Vanilla & 1.24 & 1.82 & 0.00 & 0.00 & 0.00 & \multirow{41.5}{*}{\centering \shortstack{Llama-3 \\ 8B}} & 18.14 & 16.47 & 35.44 & \textbf{15.00} & 0.00 & \multirow{41.5}{*}{\centering \shortstack{Llama-3 \\ 70B}} & 14.85 & 9.96 & 40.91 & \textbf{17.26} & \textbf{4.55}\\
                                & & {GoLLIE}& {8.26} & {9.59} & {6.90} & {5.02} & {0.00} & & {9.21} & {9.87} & {8.60} & {8.22} & {0.00} & & 11.02 & 6.84 & 36.84 & 10.96 & 0.00 \\
                                & & {GNER}& \textbf{26.98} & \textbf{30.50} & \textbf{27.35} & \textbf{16.36} & \textbf{0.44} & & {33.18} & {44.14} & {14.58} & {7.38} & {0.00} & & {24.90} & {29.05} & {27.35} & {10.00} & {1.23} \\
                                & & {CodeNER}& {17.51} & {18.18} & {27.12} & {10.43} & 0.00 & & {31.96} & {39.22} & {34.04} & 6.54 & 0.00 & & 19.39 & 18.46 & \textbf{37.25} & 13.02 & 1.13 \\
                                & & {CodeNER w/o label}& {8.49} & {7.35} & {19.05} & {6.62} & 0.00 & & \textbf{45.62} & \textbf{57.26} & \textbf{41.12} & {9.40} & \textbf{1.33} & &  \textbf{26.11} & \textbf{29.29} & 36.17 & 9.89 & 1.49 \\
\cmidrule{2-8}\cmidrule{10-14}\cmidrule{16-20}
                                & \multirow{5}{*}{\centering CoNLL03} & Vanilla & 11.73 & 21.20 & 15.44 & 3.75 & 0.00 & & \textbf{64.02} & \textbf{88.90} & \textbf{71.01} & \textbf{54.85} & 11.85 & & \textbf{68.13} & \textbf{93.39} & \textbf{74.01} & \textbf{57.40} & 21.32 \\
                                & & {GoLLIE}& {35.39} & {51.47} & {33.88} & \textbf{35.43} & {1.84} & & {34.72} & {49.03} & {38.32} & {30.65} & {2.84} & & 53.93 & 72.04 & 60.05 & 44.54 & 19.87  \\
                                & & {GNER}& {44.63} & \textbf{69.87} & {50.31} & {31.88} & {3.14} & & {34.43} & {44.81} & {49.12} & {49.12} & {1.06} & & {53.38} & {77.48} & {58.52} & {38.96} & {19.79}\\
                                & & {CodeNER}& {36.57} & {54.03} & {44.64} & {19.75} & {16.95} & & 40.29 & 51.07 & 50.70 & 31.01 & \textbf{12.71} & & 54.94 & 76.42 & 57.48 & 43.45 & 26.61 \\
                                & & {CodeNER w/o label}& \textbf{48.33} & {66.92} & \textbf{55.66} & {32.26} & \textbf{26.12} & & {37.00} & {44.27} & {46.17} & {31.80} & {10.75} & & 47.53 & 60.61 & 53.89 & 36.43 & \textbf{28.59} \\
\cmidrule{2-8}\cmidrule{10-14}\cmidrule{16-20}
                                & \multirow{5}{*}{\centering SwissNER} & Vanilla & 4.14 & 5.45 & 3.23 & 5.71 & 1.72 & & \textbf{49.04}& \textbf{72.91} & 58.82 & \textbf{52.06} & 6.15 & & \textbf{59.82} & \textbf{80.18} & \textbf{71.09} & \textbf{66.30} & 11.85 \\
                                & & {GoLLIE}& \textbf{32.54} & \textbf{51.64} & {40.51} & \textbf{32.38} & {2.42}& &{24.98} & {38.58} & {29.10} & {24.56} & {6.51} & & 44.49 & 54.78 & 62.54 & 42.62 & 9.15\\
                                & & {GNER} & {22.37} & {31.73} & {26.84} & {21.36} & {8.29} & & {25.58} & {28.27} & {38.34} & {25.74} & {2.29} & & {43.07} & {55.90} & {58.67} & {40.10} & {11.21} \\
                                & & {CodeNER} & {28.76} & {49.47} & \textbf{42.34} & {14.35} & {12.44} & & 35.48 & 28.99 & \textbf{62.37} & 30.55 & {6.55} & & 47.04 & 66.92 & 61.57 & 42.23 & \textbf{13.55} \\
                                & & {CodeNER w/o label}& {31.78} & {50.55} & {41.67} & {21.71} & \textbf{15.76} & & {31.87} & {25.37} & {54.41} & {28.09} & \textbf{8.02} & & 39.21 & 58.52 & 52.50 & 31.90 & 12.50 \\
\cmidrule{2-8}\cmidrule{10-14}\cmidrule{16-20}
                                & \multirow{5}{*}{\centering Arabic A} & Vanilla & 0.00 & 0.00 & 0.00 & 0.00 & 0.00 & & 8.88 & 6.13 & 15.53 & {6.25} & 6.31 & & {27.25} & \textbf{49.61} & {34.62} & \textbf{23.02} & {10.24}\\
                                & & {GoLLIE}& \textbf{14.60} & \textbf{16.74} & \textbf{22.80} & \textbf{8.30} & \textbf{8.76} & & {12.57} & {26.4} & {16.36} & {7.25} & {2.99} & & 22.83 & 32.84 & 35.46 & 14.14 & 9.89\\
                                & & {GNER}& {4.23} & {3.61} & {1.75} & {0.00} & {6.92} & & {6.53} & {12.09} & {11.93} & {0.00} & {0.71} & & {22.66} & {37.55} & {33.80} & {13.04} & {8.12} \\
                                & & {CodeNER} & {5.04} & {10.71} & {4.19} & 0.00 & {3.21}  & & \textbf{27.52} & \textbf{36.30}  & \textbf{34.08} & 1.59 & \textbf{22.35} & & 28.83 & 40.15 & 36.96 & 9.85 & 19.80\\
                                & & {CodeNER w/o label}& {8.81} & {13.47} & {11.76} & {2.20} & {5.21} & & {21.84} & {31.29} & {21.05} & \textbf{9.14} & {19.05} & & \textbf{29.00} & {40.61} & \textbf{37.18} & {3.98} & \textbf{20.69}\\
\cmidrule{2-8}\cmidrule{10-14}\cmidrule{16-20}
                                & \multirow{5}{*}{\centering Arabic B} & Vanilla & 0.00 & 0.00 & 0.00 & 0.00 & 0.00 & & 4.62 & 5.78 & 5.33 & \textbf{9.68} & 2.41 & & \textbf{33.68} & \textbf{62.17} & \textbf{49.29} & 14.29 & 7.35\\
                                & & {GoLLIE}& \textbf{15.83} & \textbf{23.67} & \textbf{26.40} & {6.39} & \textbf{5.58} & & {16.53} & {27.07} & {18.98} & {9.33} & {8.80} & & 25.12 & 37.79 & 34.29 & 11.65 & 12.97 \\
                                & & {GNER}& {2.14} & {3.45} & {1.23} & {2.74} & {1.64} & & {2.34} & {4.62} & {2.72} & {0.00} & {0.89} & & {24.20} & {39.53} & {44.36} & {1.63} & {5.22}\\
                                & & {CodeNER} & {6.79} & {11.76} & {2.88} & \textbf{11.43} & {4.69} & & \textbf{27.52} & {36.30} & \textbf{34.08} & 1.59 & \textbf{22.35} & & 32.16 & {46.15} & 37.25 & \textbf{17.20} & 21.78 \\
                                & & {CodeNER w/o label}& {9.09} & {13.53} & {17.07} & {0.00} &  {2.92} & & {24.69} & \textbf{36.31} & {31.15} & {5.85} & {16.18} & & {33.62} & {45.52} & {40.14} & {15.66} & \textbf{24.45}\\
\cmidrule{2-8}\cmidrule{10-14}\cmidrule{16-20}
                                & \multirow{5}{*}{\centering Finnish A} & Vanilla & 1.92 & 1.08 & 0.00 & 3.95 & 0.00 & & 23.80 & \textbf{42.90} & 20.74 & 27.99 & 1.71 & & {39.78} & \textbf{56.03} & {52.20} & {44.30} & {0.87} \\
                                & & {GoLLIE}& {23.48} & {28.22} & {16.89} & \textbf{33.52} & {3.24} & & {21.86} & {22.05} & {17.80} & {31.99} & {2.15} & & 34.15 & 43.23 & 44.65 & 38.38 & 4.07\\
                                & & {GNER}& {18.40} & {13.64} & {24.73} & {23.63} & {2.20} & & {7.50} & {9.49} & {9.88} & {8.29} & {1.13} & & {34.80} & {44.96} & {40.57} & {40.71} & {5.44}\\
                                & & {CodeNER} & {27.34} & {39.63} & {34.71} & {27.44} & \textbf{8.28} & & {29.46} & 31.27 & {42.03} & {29.02} & \textbf{14.11} & & 48.89 & {45.72} & 70.74 & 50.94 & \textbf{20.63}\\
                                & & {CodeNER w/o label}& \textbf{31.69} & \textbf{42.40} & \textbf{44.13} & {32.61} & {5.95} & & \textbf{32.11} & {21.01} & \textbf{45.98} & \textbf{39.24} & {7.58} & & \textbf{48.98} & {43.44} & \textbf{72.01} & \textbf{51.63} & {19.81}\\
\cmidrule{2-8}\cmidrule{10-14}\cmidrule{16-20}
                                & \multirow{5}{*}{\centering DaNE} & Vanilla & 0.00 & 0.00 & 0.00 & 0.00 & 0.00 & & 13.49 & 17.35 & 9.43 & 18.09 & 4.84 & & {13.50} & {17.82} & {11.54} & {20.00} & {0.00}\\
                                & & {GoLLIE}& {27.37} & {44.23} & {25.95} & {26.28} & {4.85} & & {25.33} & {38.40} & {25.20} & {25.71} & {5.48} & & \textbf{47.17} & \textbf{64.00} & 56.35 & \textbf{46.87} & 15.23 \\
                                & & {GNER}& {31.42} & {56.27} & {14.36} & {30.86} & {8.75} & & {26.92} & {45.89} & {22.15} & {27.86} & {1.26} & & {39.15} & {44.37} & {60.82} & {47.65} & {2.86}\\
                                & & {CodeNER} & {33.79} & {54.15} & {53.59} & {19.72} & {6.52} & & \textbf{37.55} & \textbf{42.29} & \textbf{59.05} & \textbf{36.13} & \textbf{15.33} & & 42.93 & {50.80} & \textbf{63.33} & 39.90 & \textbf{19.08} \\
                                & & {CodeNER w/o label}& \textbf{41.27} & \textbf{61.10} & \textbf{58.33} & \textbf{30.48} & \textbf{12.57} & & {30.02} & {32.61} & {49.56} & {34.11} & {5.22} &  & {35.06} & {38.22} & {54.81} & {33.54} & {16.74}\\
\cmidrule{2-8}\cmidrule{10-14}\cmidrule{16-20}
                                & \multirow{5}{*}{\centering Average} & Vanilla & 2.72 & 4.22 & 2.67 & 1.92 & 0.25 & & 26.00 & 35.78 & 30.90 & \textbf{26.27} & 4.75 & & {36.72} & \textbf{52.74} & {47.67} & \textbf{34.65} & {8.03}\\
                                & & {GoLLIE}& {22.50} & {32.22} & {24.76} & \textbf{21.05} & {3.81} & & {20.74} & {30.2} & {22.05} & {19.67} & {4.11} & & {34.10} & {44.50} & 47.17 & {29.88} & 10.17 \\
                                & & {GNER} & {21.45} & {29.87} & {20.94} & {18.12} & {4.48} & & {19.50} & {27.04} & {21.25} & {13.28} & {1.05} & & {34.59} & {46.98} & {46.30} & {27.44} & {7.70} \\
                                & & {CodeNER} & {22.26} & {33.99} & {29.92} & {14.73} & {7.44} & & {31.81} & \textbf{37.74} & \textbf{43.38} & 19.44 & 
                                \textbf{12.23} & & \textbf{39.17} & {49.23} & \textbf{52.08} & 30.94 & {17.51}\\
                                & & {CodeNER w/o label}& \textbf{25.64} & \textbf{36.47} & \textbf{35.38} & {17.98} & \textbf{9.79} & & \textbf{31.88} & {35.45} & {41.35} & {22.52} & {9.73} &  & {37.07} & {45.17} & {49.53} & {26.15} & \textbf{17.75} \\
\bottomrule
\end{tabular}}
% \end{adjustbox}
\caption{Experimental results for each label on \texttt{FIN}, \texttt{CoNLL03}, and \texttt{SwissNER} using the Phi- and Llama-3 models. CodeNER w/o label indicates that we removed entity-label descriptions from the prompt.}
\label{tab:Phiresults}
\end{table*}

\section{Analysis}
\subsection{Open Models}
We evaluated the proposed CodeNER with the open models, Phi-3 and Llama-3. 
We additionally compared our CodeNER with the prompts from zero-shot NER models, GoLLIE and GNER, which are partial code-based prompts that focus on context information and BIO output schema without code, respectively~\cite{sainz2024gollie,ding-etal-2024-rethinking}. %\footnote{GoLLIE prompt is described in Appendix~\ref{appen:GoLLIE}.}
Table~\ref{tab:Phiresults} shows the $F_1$ scores, including the detailed performance for each label. For both models, the proposed CodeNER consistently outperformed Vanilla with an average improvement of 19.54, 5.81, and 2.45  points in the $F_1$ scores.%, representing 718.4\% and 34.22\% improvements, respectively. 

Significant differences were observed between the Phi-3 and Llama-3-8B models. We think that the differences in performance between Phi-3 and Llama-3-8B are closely related to the capability of the models to interpret programming language instructions. While Llama-3-8B appears to have a better understanding of the programming language instructions on \texttt{FIN} and \texttt{Arabic}, particularly for NER tasks that require long-range scopes, Phi-3 may require fine-tuning approaches for further improvement. 
Including entity-label descriptions is not necessary because the averaged scores of CodeNER and CodeNER w/o label exhibit similar performance. 

CodeNER significantly outperformed GoLLIE and GNER prompts. By employing a structured code-based approach, CodeNER leverages both BIO output schema and input sentence information, demonstrating that leveraging code is important for improved performance. We also conducted experiments with an increased model size using Llama-3-70B. Our CodeNER outperformed the current prompts from zero-shot NER models of GoLLIE and GNER. 

The results presented in Table~\ref{tab:result_label_vanilla}, including prompts with curated descriptions and vanilla outputs, highlight the effectiveness of our CodeNER.

\subsection{Impact of BIO Schema}
To investigate the effectiveness of incorporating BIO schema within code-based prompts, we evaluated only the \quotes{B} and \quotes{O} tags, which are crucial for accurately identifying entity span boundaries. Table~\ref{tab:BOtags} shows the results. Our CodeNER demonstrates the importance of integrating both the BIO output schema and broader context information of input. %By incorporating \textcolor{red}{BIOE schema within code-based prompts}, 
CodeNER outperformed the text-based prompt approach of Vanilla. Furthermore, CodeNER surpassed the code-based prompt method of GoLLIE by effectively recognizing start and end tokens rather than relying solely on input sentence information.

\begin{table}[t!]
\renewcommand{\arraystretch}{0.8}

% \begin{adjustbox}{width=0.95\columnwidth,center}
\resizebox{0.5\textwidth}{!}{

\centering
\huge
\begin{tabular}{ccccccccc}
\toprule
\rowcolor{gray!10}
\textbf{Model} &\textbf{Dataset} &\textbf{Template} &{\textbf{$F_1$}} & {\textbf{PER}} & {\textbf{LOC}} &{\textbf{ORG}} & {\textbf{MISC}} & {\textbf{O}}\\ 
\midrule
\multirow{21}{*}{\centering Phi-3} & \multirow{5}{*}{\centering FIN} 
                                & Vanilla & 23.55 & 4.55 & 4.88 & 10.00 & 0.00 & \textbf{98.33} \\
                                & & {GoLLIE}& {25.86} & {11.81} & {16.09} & {9.40} & {0.00} & {92.02}\\
                                & & {GNER}& \textbf{36.66} & \textbf{32.62} & \textbf{32.48} & \textbf{22.30} & \textbf{0.44} & {95.46}\\
                                & & {CodeNER}& {32.71} & {18.25} & {34.48} & {13.10} & 0.00 & {97.71} \\
                                & & {CodeNER w/o label}& {31.13} & {13.11} & {25.81} & {18.67} & 0.00 & {98.05} \\
\cmidrule{2-9}
                                & \multirow{5}{*}{\centering CoNLL03} & Vanilla & 27.34 & 22.86 & 16.20 & 6.14 & 0.51 & {91.00} \\
                                & & {GoLLIE}& {43.54} & {52.38} & {37.96} & {38.10} & {3.08} & {86.20}\\
                                & & {GNER}& {51.28} & \textbf{72.62} & {52.07} & \textbf{38.34} & {7.42} & {85.96}\\
                                & & {CodeNER} & {51.26} & {58.36} & {47.97} & {22.90} & {33.28} & {93.78} \\
                                & & {CodeNER w/o label}& \textbf{58.48} & {70.34} & \textbf{56.60} & {36.84} & \textbf{33.63} & \textbf{94.96} \\
\cmidrule{2-9}
                                & \multirow{5}{*}{\begin{tabular}{@{}c@{}}SwissNER \\ Arabic A \& B \\ Finnish A \\ DaNE\end{tabular}} & Vanilla & {19.95} & {1.31} & {0.86} & {3.08} & {0.34} & {94.16} \\
                                & & {GoLLIE}& {37.00} & {35.30} & {30.55} &  \textbf{24.55} & {6.86} & {87.72} \\
                                & & {GNER}& {28.95} & {16.37} & {14.90} & {14.15} & {5.50} & {93.85}\\
                                & & {CodeNER} & {38.38} & {38.44} & {29.63} & {18.49} & {10.57} & {94.76} \\
                                & & {CodeNER w/o label}& \textbf{41.03} & \textbf{40.08} & \textbf{35.59} & {22.46} & \textbf{12.23} & \textbf{94.78}\\
\cmidrule{2-9}

                                & \multirow{5}{*}{\centering Average} & Vanilla & 21.52 & 4.85 & 3.63 & 4.50 & 0.32 & 94.30 \\
                                & & {GoLLIE}& {36.34} & {34.38} & {29.54} & \textbf{24.32} & {5.34} & {88.12} \\
                                & & {GNER}& {33.96} & {28.45} & {24.02} & {19.54} & {4.97} & {92.80}\\
                                & & {CodeNER} & {39.41} & {38.40} & {32.94} & {18.35} & {12.30} & {95.04} \\
                                & & {CodeNER w/o label}& \textbf{42.11} & \textbf{40.55} & \textbf{37.19} & {23.97} & \textbf{13.54} & \textbf{95.27} \\
\midrule
\multirow{21}{*}{\centering Llama-3} & \multirow{5}{*}{\centering FIN} & Vanilla & 36.79 & 19.61 & {43.04} & \textbf{20.83} & {3.23} & {97.22} \\
                                   & & {GoLLIE}& {27.51} & {11.84} & {19.35} & {10.96} & {0.00} & {95.40} \\
                                   & & {GNER}& {33.66} & {45.78} & {14.58} & {10.33} & {0.00} & \textbf{97.62}\\
                                   & & {CodeNER} & {36.91} & {43.82} & {34.41} & 8.33 & 1.80 & {96.20} \\
                                   & & {CodeNER w/o label}& \textbf{42.89} & \textbf{59.88} & \textbf{43.40} & {12.32} & \textbf{3.33} & {95.55} \\
\cmidrule{2-9}
                                   & \multirow{5}{*}{\centering CoNLL03} & Vanilla & \textbf{65.82} & \textbf{89.09} & \textbf{71.81} & \textbf{56.20} & 15.08 & \textbf{96.91} \\
                                   & & {GoLLIE}& {43.28} & {49.48} & {40.62} & {32.63} & {4.13} & {89.54} \\
                                   & & {GNER}& {46.16} & {52.56} & {51.45} & {31.30} & {1.91} & {93.56}\\
                                   & & {CodeNER} & 51.62 & 54.81 & 52.80 & 38.66 & \textbf{17.82} & {94.02} \\
                                   & & {CodeNER w/o label}& {49.45} & {49.77} & {49.21} & {40.02} & {14.91} & {93.32} \\
\cmidrule{2-9}
                                   & \multirow{5}{*}{\begin{tabular}{@{}c@{}}SwissNER \\ Arabic A \& B \\ Finnish A \\ DaNE\end{tabular}} & Vanilla & {35.53} & {29.25} & {23.32} & {25.56} & {4.72} & \textbf{94.80} \\
                                   & & {GoLLIE} & {34.87}& {31.08} & {23.50} & {22.48} & {6.70} & {90.61} \\
                                   & & {GNER}& {27.96} & {15.42} & {17.16} & {13.59} & {1.94} & {91.69}\\
                                   & & {CodeNER} & \textbf{45.46}& \textbf{41.67} & \textbf{47.31} & {27.10} & \textbf{16.89} & {94.32} \\
                                   & & {CodeNER w/o label} & {42.93}& {34.89} & {43.70} & \textbf{29.49} & {13.44} & {93.13} \\
\cmidrule{2-9}
                                & \multirow{5}{*}{\centering Average} & Vanilla & 40.04 & 36.42 & 33.06 & \textbf{29.26} & 5.99 & \textbf{95.45} \\
                                & & {GoLLIE}& {35.02} & {30.96} & {25.35} & {22.28} & {5.38} & {91.14} \\
                                & & {GNER}& {31.94} & {26.67} & {22.44} & {16.00} & {1.61} & {92.99}\\
                                & & {CodeNER} & \textbf{45.12} & \textbf{43.86} & \textbf{46.25} & 26.07 & 
                                \textbf{14.87} & {94.54} \\
                                & & {CodeNER w/o label}& {43.86} & {40.59} & {44.45} & {28.54} & {12.20} & {93.50} \\
\bottomrule
\end{tabular}
}
% \end{adjustbox}
\caption{Experimental results for only \quotes{B} and \quotes{O} tags for each label using the Phi- and Llama-3 models. For \texttt{SwissNER}, \texttt{Arabic}, \texttt{Finnish}, and \texttt{DaNE}, we averaged their scores.} %on \texttt{FIN}, \texttt{CoNLL03}, and \texttt{SwissNER} using the Phi- and Llama-3 models.}
\label{tab:BOtags}
\end{table}

\begin{figure*}[ht!]
  \centering
  \includegraphics[width=0.85\linewidth]{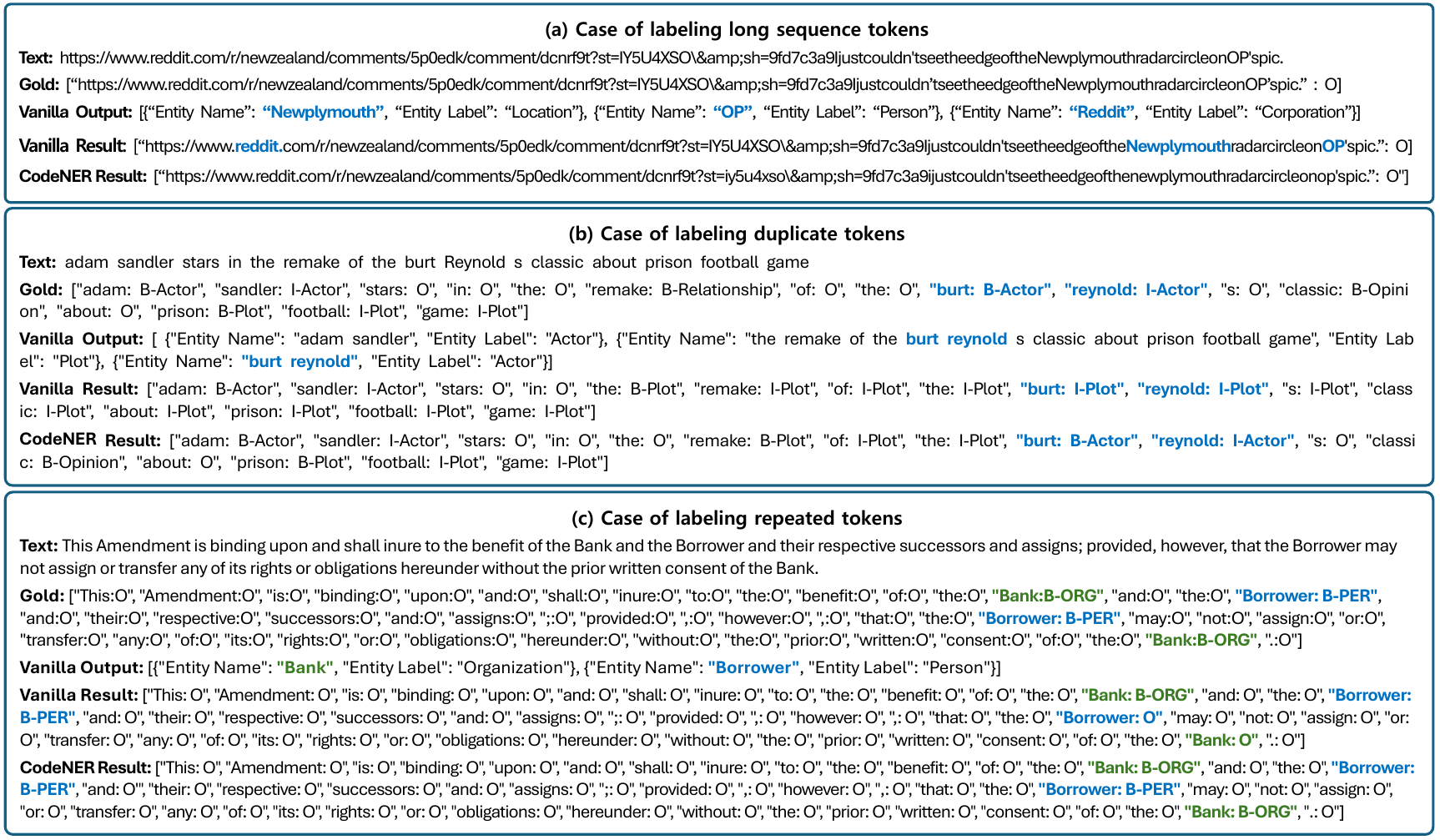}
  \includegraphics[width=0.85 \linewidth]{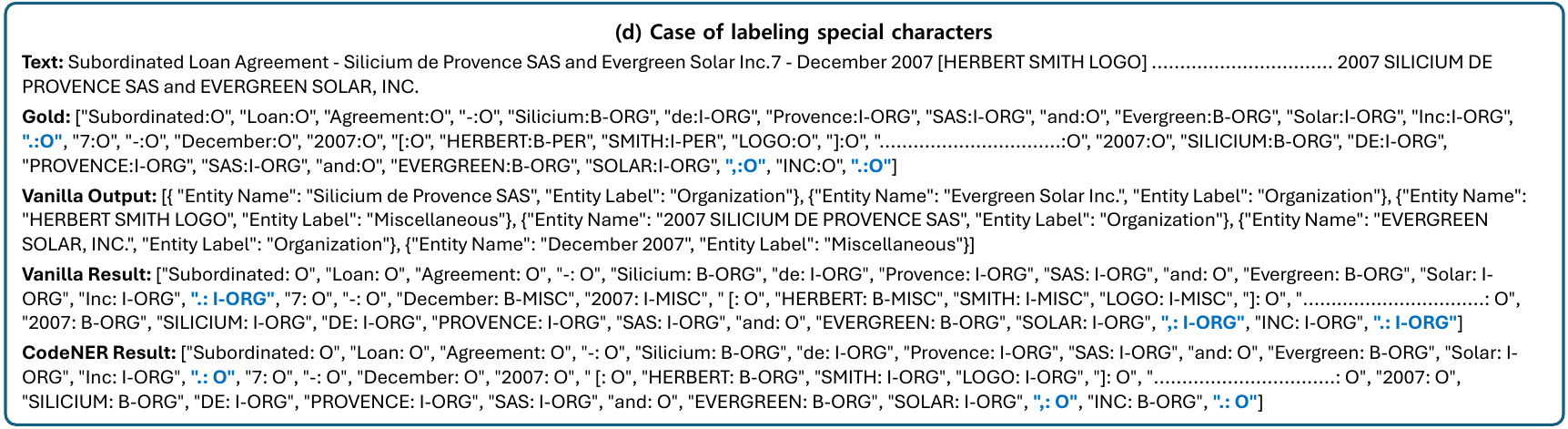}
  \caption{Case studies of Vanilla and CodeNER outputs. \textbf{Text} denotes the input sentence, and \textbf{Gold} refers to the correct label for \textbf{Text}. \textbf{Vanilla Output} denotes the results generated by Vanilla.  \textbf{Vanilla Result} represents the conversion of \textbf{Vanilla Output} into  the BIO-tagging format for evaluation. \textbf{CodeNER Result} denotes the CodeNER labeling results for \textbf{Text}.}
  \label{fig:caseStudy}
\end{figure*}

\subsection{Case Study}

We conducted a case study of the outputs from CodeNER and Vanilla. The results are shown in Figure~\ref{fig:caseStudy}. In Figure~\ref{fig:caseStudy} (a), Vanilla failed to treat the website as a single token; instead, it extracted and labeled individual words. In contrast, the proposed CodeNER accurately recognized the website as a single token. In Figure~\ref{fig:caseStudy} (b), while Vanilla generated duplicate labels for the same word, CodeNER did not suffer from this problem because it can capture the structured token-by-token approach. In Figure~\ref{fig:caseStudy} (c), when a word is repeated multiple times in the text, Vanilla tended to annotate labels only once, whereas CodeNER did not. In Figure~\ref{fig:caseStudy} (d), Vanilla could not capture special tokens, such as comma and period, when attached to a word because it relies solely on sentence-level information. However, CodeNER successfully identified special tokens. Details of evaluations for such cases are described in Appendix~\ref{appen:dup}.

\subsection{Combination with Chain-of-thought}
We evaluated the proposed CodeNER by incorporating the CoT reasoning instruction, \quotes{Let’s think step by step}, at the beginning of the CodeNER instruction. Although previous text-based vanilla prompting methods have been also explored to incorporate CoT in NER~\cite{han2023informationextractionsolvedchatgpt,xie-etal-2023-empirical}, the proposed approach, CodeNER, presents a different perspective. It encourages LLMs to enhance their understanding of long-range scopes using programming language-style prompts. Table~\ref{tab:COTresults} shows the results. The integration of CoT with CodeNER demonstrated the potential to improve performance in zero-shot settings by leveraging structured programming language-style prompts.

\subsection{Different Programming Languages}
It is also of interest to perform a comparison of different programming languages such as C++.\footnote{CodeNER with C++ prompts is in Appendix~\ref{appen:c++}.} Table~\ref{tab:cppresults} shows the results. CodeNER with C++ instructions showed significant improvements in the $F_1$ scores over both Vanilla and CodeNER with Python instructions on \texttt{FIN}. Although the consideration of C++ showed better person entity recognition performance on \texttt{FIN}, the differences between the two programming languages in prompts were less noticeable.

Interestingly, the $F_1$ scores with CodeNER remain consistent between the Python and C++ instructions on \texttt{CoNLL03}. We think this suggests that dataset characteristics may have a more significant impact than the choice of the programming language. The comparable results between Python and C++ suggest that the choice of the programming language has a minimal impact on the effectiveness of CodeNER when used on datasets where the model's existing capabilities are strong. We additionally considered using the Java programming language for CodeNER. However, the performance was nearly zero in terms of the F$_1$ scores, which indicates LLMs lack inherent knowledge of Java in their pre-training~\cite{haroon2025accuratelylargelanguagemodels}.

\begin{table}[t!]
% \begin{adjustbox}{width=.9\columnwidth,center}
\renewcommand{\arraystretch}{0.8}
\centering
\huge
\resizebox{0.5\textwidth}{!}{
\begin{tabular}{cccccccc}
\rowcolor{gray!10}
\toprule
\textbf{Model} &\textbf{Dataset} &\textbf{Instruction} &{\textbf{$F_1$}} & {\textbf{PER}} & {\textbf{LOC}} & {\textbf{ORG}} & {\textbf{MISC}} \\
\midrule
\multirow{15}{*}{\centering GPT-4} & \multirow{7}{*}{\centering FIN}& 
Vanilla       & 22.83 & 26.94 & 15.62 & 14.72 & {1.06} \\
              &           & 
Vanilla+CoT       & 24.29 & 29.27 & 11.94 & \textbf{16.59} & 0.91 \\
\cmidrule{3-8}
              &           &
GoLLIE        & {12.59} & {8.75} & \textbf{44.16} & {7.14} & \textbf{1.91} \\
              &           & 
GoLLIE+CoT    & {12.52} & {8.93} & {42.67} & {7.21} & {0.89} \\
\cmidrule{3-8}
            &           & 
CodeNER       & 41.13 & 51.71 & 38.10 & 7.48 & 0.78 \\
            &           & 
CodeNER+CoT    & \textbf{42.09} & \textbf{52.69} & {38.60} & 8.80  & 0.87 \\
\cmidrule{2-8}
              & \multirow{7}{*}{\centering CoNLL03} & 
Vanilla & \textbf{73.24} & \textbf{92.06} & 79.45 & \textbf{66.15} & 31.88 \\
            &           & 
Vanilla+CoT & 72.53 & 91.71 & \textbf{79.61} & 64.64 & 30.18 \\
\cmidrule{3-8}
            &           &
GoLLIE      & {51.96} & {65.97} & {62.48} & {46.89} & {6.67} \\
            &           & 
GoLLIE+CoT  & {51.76} & {65.62} & {62.22} & {46.85} & {6.57} \\
\cmidrule{3-8}
            &           & 
CodeNER & 69.48 & 88.38 & 75.77 & 58.53 & 36.87 \\
            &           & 
CodeNER+CoT & 69.32 & 87.57 & 76.47 & 57.86 & \textbf{37.38} \\

\midrule
\multirow{15}{*}{\centering Phi-3} & \multirow{7}{*}{\centering FIN}& 
Vanilla       & 1.24 & 1.82 & 0.00 & 0.00 & 0.00 \\
              &           & 
Vanilla+CoT       & 1.84 & 2.70 & 0.00 & 0.00 & 0.00 \\
\cmidrule{3-8}
              &           &
GoLLIE        & {8.26} & {9.59} & {6.90} & {5.02} & {0.00} \\
              &           & 
GoLLIE+CoT    & {4.66} & {4.27} & {7.23} & {4.82} & \textbf{1.11} \\
\cmidrule{3-8}
            &           & 
CodeNER       & 17.51 & 18.18 & 27.12 & 10.43 & 0.00 \\
            &           & 
CodeNER+CoT    & \textbf{18.73} & \textbf{19.61} & \textbf{28.57} & \textbf{10.86} & 0.00 \\
\cmidrule{2-8}
              & \multirow{7}{*}{\centering CoNLL03} & 
Vanilla & 11.73 & 21.20 & 15.44 & 3.75 & 0.00 \\
            &           & 
Vanilla+CoT & 12.75 & 22.27 & 16.43 & 4.96 & 0.53 \\
\cmidrule{3-8}
            &           &
GoLLIE      & {35.39} & {51.47} & {33.88} & \textbf{35.43} & {1.84} \\
            &           & 
GoLLIE+CoT  & {27.13} & {37.23} & {27.37} & {27.68} & {1.97} \\
\cmidrule{3-8}
            &           & 
CodeNER & 36.57 & 54.03 & 44.64 & 19.75 & 16.95 \\
            &           & 
CodeNER+CoT & \textbf{44.62} & \textbf{61.94} & \textbf{54.63} & {26.98} & \textbf{22.66} \\

\midrule
\multirow{15}{*}{\centering Llama-3} & \multirow{7}{*}{\centering FIN}& 
Vanilla       & 18.14 & 16.47 & \textbf{35.44} & 15.00 & 0.00 \\
              &           & 
Vanilla+CoT       & 18.47 & 16.34 & 37.50 & \textbf{15.77} & {1.56} \\
\cmidrule{3-8}
            &           &
GoLLIE      & {9.21} & {9.87} & {8.60} & {8.22} & {0.00} \\
            &           & 
GoLLIE+CoT  & {9.92} & {10.29} & {14.29} & {6.14} & \textbf{4.49} \\
\cmidrule{3-8}
            &           & 
CodeNER       & \textbf{31.96} & \textbf{39.22} & 34.04 & 6.54 & 0.00 \\
            &           & 
CodeNER+CoT    & 27.79 & 35.15 & 22.22 & 6.77 & 0.00 \\
\cmidrule{2-8}
              & \multirow{7}{*}{\centering CoNLL03} & 
Vanilla & 64.02 & \textbf{88.90} & \textbf{71.01} & \textbf{54.85} & 11.85 \\
            &           & 
Vanilla+CoT & \textbf{64.08} & {88.70} & {70.70} & {54.52} & 14.26 \\
\cmidrule{3-8}
            &           &
GoLLIE      & {34.72} & {49.03} & {38.32} & {30.65} & {2.84} \\
            &           & 
GoLLIE+CoT  & {38.53} & {52.08} & {43.78} & {35.23} & {2.67} \\
\cmidrule{3-8}
            &           & 
CodeNER & 40.29 & 51.07 & 50.70 & 31.01 & 12.71 \\
            &           & 
CodeNER+CoT & 44.21 & 55.68 & 56.68 & 32.32 & \textbf{16.29} \\

\bottomrule
\end{tabular}}

% \end{adjustbox}

\caption{Experimental results including each label on \texttt{FIN} and \texttt{CoNLL03} by incorporating CoT.}
\label{tab:COTresults}
\end{table}

\begin{table}[t!]
% \begin{adjustbox}{width=.95\columnwidth,center}
\renewcommand{\arraystretch}{0.8}
\centering
\huge
\resizebox{0.5\textwidth}{!}{
\begin{tabular}{cccccccc}
\rowcolor{gray!10}
\toprule
\textbf{Model} &\textbf{Dataset} &\textbf{Instruction} &{\textbf{$F_1$}} & {\textbf{PER}} & {\textbf{LOC}} & {\textbf{ORG}} & {\textbf{MISC}} \\
\midrule
\multirow{8.5}{*}{\centering GPT-4} & \multirow{4}{*}{\centering FIN} & Vanilla & 22.83 & 26.94 & 15.62 & \textbf{14.72} & {1.06} \\
            & & GoLLIE & {12.59} & {8.75} & \textbf{44.16} & {7.14} & \textbf{1.91} \\
            & & CodeNER (Python)    & 41.13 & 51.71 & {38.10} &  7.48 & 0.78 \\
            & & CodeNER (C++) & \textbf{44.44} & \textbf{56.84} & 33.33 &  9.85 & 0.55 \\
\cmidrule{2-8}
            & \multirow{4}{*}{\centering CoNLL03} & Vanilla       & \textbf{73.24} & \textbf{92.06} & \textbf{79.45} & \textbf{66.15} & 31.88 \\
            & & GoLLIE & {51.96} & {65.97} & {62.48} & {46.89} & {6.67} \\
            & & CodeNER (Python)    & 69.48 & 88.38 & 75.77 & 58.53 & 36.87  \\
            & & CodeNER (C++) & 68.43 & 85.98 & 74.44 & 58.59 & \textbf{37.01}  \\
\midrule       
\multirow{8.5}{*}{\centering Phi-3} & \multirow{4}{*}{\centering FIN} & Vanilla & 1.24 & 1.82 & 0.00 & 0.00 & 0.00 \\
            & & GoLLIE & {8.26} & {9.59} & {6.90} & {5.02} & {0.00} \\
            & & CodeNER (Python)    & \textbf{17.51} & \textbf{18.18} & \textbf{27.12} & 10.43 & 0.00 \\
            & & CodeNER (C++) & 14.61 & 14.05 & 12.00 & \textbf{20.41} & 0.00 \\
\cmidrule{2-8}
            & \multirow{4}{*}{\centering CoNLL03} & Vanilla       & 11.73 & 21.20 & 15.44 & 3.75 & 0.00 \\
            & & GoLLIE & {35.39} & {51.47} & {33.88} & \textbf{35.43} & {1.84} \\
            & & CodeNER (Python)    & 36.57 & 54.03 & 44.64 & 19.75 & \textbf{16.95} \\
            & & CodeNER (C++) & \textbf{38.44} & \textbf{56.84} & \textbf{47.63} & {21.57} & 14.15 \\
\midrule            
\multirow{8.5}{*}{\centering Llama-3} & \multirow{4}{*}{\centering FIN} & Vanilla & 18.14 & 16.47 & \textbf{35.44} & \textbf{15.00} & 0.00 \\
            & & GoLLIE & {9.21} & {9.87} & {8.60} & {8.22} & {0.00} \\
            & & CodeNER (Python) & \textbf{31.96} & \textbf{39.22} & {34.04} & 6.54 & 0.00 \\
            & & CodeNER (C++) & 25.08 & 31.62 & 19.23 & {7.07} & 0.00 \\
\cmidrule{2-8}
            & \multirow{4}{*}{\centering CoNLL03} & Vanilla       & \textbf{64.02} & \textbf{88.90} & \textbf{71.01} & \textbf{54.85} & 11.85 \\
            & & GoLLIE & {34.72} & {49.03} & {38.32} & {30.65} & {2.84} \\
            & & CodeNER (Python)    & 40.29 & 51.07 & 50.70 & 31.01 & 12.71 \\
            & & CodeNER (C++) & 32.84 & 42.23 & 44.05 & 20.28 & \textbf{14.25} \\
\bottomrule
\end{tabular}
}
% \end{adjustbox}

\caption{Experimental results including each label on \texttt{FIN} and \texttt{CoNLL03} with different programming languages.}
\label{tab:cppresults}
\end{table}

\subsection{Trade-off Discussion}
\noindent \textbf{Advantages.}
The proposed CodeNER %, which considers both token- and sentence-level information for NER, 
provides several advantages because it successfully bridges the gap between the text-in-text-out schema of LLMs and the text-in-span-out nature for handling the sequential aspect of BIO output schema.
The Vanilla prompt method, during labeling, leads to overlapping labels, which results in unclear classifications; in contrast, CodeNER leverages sequential token information through a for-loop, which avoids errors from overlapping labels and ensures that each token is labeled only once. 
In cases requiring the identification of individual tokens, CodeNER accurately recognizes a token as a single unit, whereas the Vanilla prompt method struggles to label tokens such as web addresses. Therefore, for datasets requiring NER to recognize single labels, CodeNER is more effective. In addition, when the same word appears multiple times, the Vanilla prompt faces difficulties in determining which tokens to label. However, CodeNER accurately labels each token. 

\noindent \textbf{Disadvantages.}
Although CodeNER effectively recognizes named entities, there are cases in which considering the BIO output schema within code blocks may not be advantageous. In our in-depth analysis of two datasets, \texttt{MIT-Movie} and \texttt{MIT-restaurant}, we found that \texttt{MIT-Movie} contains long sentences in which the labels \quotes{Plot} and \quotes{Quote} appear frequently. These labels occupy significant portions (42.61\%) of the dataset. They typically encompass entire sentences or long text spans that include function words such as conjunctions, prepositions, and determiners, for example, \quotes{in,} \quotes{is,} \quotes{with,} and \quotes{the}, while such words do not convey specific meanings. Consequently, for the datasets where sentences or long text spans with function words need to be labeled, the Vanilla prompt method, which considers only input sentence information, can be more effective than CodeNER.\footnote{Additional case studies are described in Appendix~\ref{appen:dis}.}

\section{Conclusion}
In this study, we introduced CodeNER, a novel code-based prompting method designed to enhance the capabilities of LLMs in performing NER. By embedding structured code within prompts, CodeNER effectively integrates the BIO output schema and input sentence information, thereby leveraging the inherent strengths of programming language comprehension in LLMs. Experimental results on ten NER benchmarks demonstrated that CodeNER consistently outperforms text-based prompting methods by bridging the inherent gap between NER and LLMs. 

\section*{Limitations}
Although CodeNER showed significant improvements by leveraging code-based prompting for NER, it relies on the capabilities of LLMs. LLMs without being trained in programming languages may require further fine-tuning. Moreover, for datasets containing long sentences or long text spans with function words, the Vanilla prompt method that considers only input sentence information can sometimes be more effective than our CodeNER. In the future, we plan to pre-train our CodeNER by incorporating a related auxiliary task that focuses on identifying function words or syntactic structures across various datasets.

% Entries for the entire Anthology, followed by custom entries
\bibliography{custom}

@phdthesis{10.5555/AAI28114605,
author = {Chen, Danqi},
advisor = {Christopher, Manning, and Dan, Jurafsky, and Percy, Liang, and Luke, Zettlemoyer,},
title = {Neural Reading Comprehension and Beyond},
year = {2018},
isbn = {9798662538269},
publisher = {Stanford University},
address = {Stanford, CA, USA},
note = {AAI28114605}
}

@misc{haroon2025accuratelylargelanguagemodels,
      title={How Accurately Do Large Language Models Understand Code?}, 
      author={Sabaat Haroon and Ahmad Faraz Khan and Ahmad Humayun and Waris Gill and Abdul Haddi Amjad and Ali R. Butt and Mohammad Taha Khan and Muhammad Ali Gulzar},
      year={2025},
      eprint={2504.04372},
      archivePrefix={arXiv},
      primaryClass={cs.SE},
      url={https://arxiv.org/abs/2504.04372}, 
}

@inproceedings{ding-etal-2024-rethinking,
    title = "Rethinking Negative Instances for Generative Named Entity Recognition",
    author = "Ding, Yuyang  and
      Li, Juntao  and
      Wang, Pinzheng  and
      Tang, Zecheng  and
      Bowen, Yan  and
      Zhang, Min",
    editor = "Ku, Lun-Wei  and
      Martins, Andre  and
      Srikumar, Vivek",
    booktitle = "Findings of the Association for Computational Linguistics: ACL 2024",
    month = aug,
    year = "2024",
    address = "Bangkok, Thailand",
    publisher = "Association for Computational Linguistics",
    url = "https://aclanthology.org/2024.findings-acl.206/",
    doi = "10.18653/v1/2024.findings-acl.206",
    pages = "3461--3475"
}

@misc{zamai2024lessinstructmoreenriching,
      title={Show Less, Instruct More: Enriching Prompts with Definitions and Guidelines for Zero-Shot NER}, 
      author={Andrew Zamai and Andrea Zugarini and Leonardo Rigutini and Marco Ernandes and Marco Maggini},
      year={2024},
      eprint={2407.01272},
      archivePrefix={arXiv},
      primaryClass={cs.CL},
      url={https://arxiv.org/abs/2407.01272}, 
}

@inproceedings{tan2021sequencetoset,
    title={A Sequence-to-Set Network for Nested Named Entity Recognition}, 
    author={Zeqi Tan and Yongliang Shen and Shuai Zhang and Weiming Lu and Yueting Zhuang},
    url = {https://arxiv.org/abs/2105.08901},
    booktitle = {Proceedings of the 30th International Joint Conference on
                 Artificial Intelligence, {IJCAI-21}},
    year = {2021},
}

@article{neuips2022,
  author       = {Long Ouyang and Jeff Wu and Xu Jiang and Diogo Almeida and Carroll L. Wainwright and Pamela Mishkin and Chong Zhang and Sandhini Agarwal and Katarina Slama and Alex Ray and John Schulman and Jacob Hilton and Fraser Kelton and Luke Miller and Maddie Simens and Amanda Askell and Peter Welinder and Paul Christiano and Jan Leike and Ryan Lowe},
  journal      = {Advances in Neural
Information Processing Systems},
  volume       = {35:27730–27744.},
  title        = {Training language models to follow instructions with human feedback},
  year         = {2022},
  eprinttype    = {Advances in Neural
Information Processing Systems},
  eprint       = {35},
}

@inproceedings{weifinetuned,
  title={Finetuned Language Models are Zero-Shot Learners},
  author={Wei, Jason and Bosma, Maarten and Zhao, Vincent and Guu, Kelvin and Yu, Adams Wei and Lester, Brian and Du, Nan and Dai, Andrew M and Le, Quoc V},
  booktitle={International Conference on Learning Representations},
  year         = {2022},
}

@incollection{51115,title	= {LaMDA: Language Models for Dialog Applications},author	= {Aaron Daniel Cohen and Adam Roberts and Alejandra Molina and Alena Butryna and Alicia Jin and Apoorv Kulshreshtha and Ben Hutchinson and Ben Zevenbergen and Blaise Hilary Aguera-Arcas and Chung-ching Chang and Claire Cui and Cosmo Du and Daniel De Freitas Adiwardana and Dehao Chen and Dmitry (Dima) Lepikhin and Ed H. Chi and Erin Hoffman-John and Heng-Tze Cheng and Hongrae Lee and Igor Krivokon and James Qin and Jamie Hall and Joe Fenton and Johnny Soraker and Kathy Meier-Hellstern and Kristen Olson and Lora Mois Aroyo and Maarten Paul Bosma and Marc Joseph Pickett and Marcelo Amorim Menegali and Marian Croak and Mark Díaz and Matthew Lamm and Maxim Krikun and Meredith Ringel Morris and Noam Shazeer and Quoc V. Le and Rachel Bernstein and Ravi Rajakumar and Ray Kurzweil and Romal Thoppilan and Steven Zheng and Taylor Bos and Toju Duke and Tulsee Doshi and Vincent Y. Zhao and Vinodkumar Prabhakaran and Will Rusch and YaGuang Li and Yanping Huang and Yanqi Zhou and Yuanzhong Xu and Zhifeng Chen},year	= {2022},booktitle	= {arXiv}}

@article{llama3modelcard,

title={Llama 3 Model Card},

author={AI@Meta},

year={2024},

url = {https://github.com/meta-llama/llama3/blob/main/MODEL_CARD.md}

}

@misc{abdin2024phi3technicalreporthighly,
      title={Phi-3 Technical Report: A Highly Capable Language Model Locally on Your Phone}, 
      author={Marah Abdin and Sam Ade Jacobs and Ammar Ahmad Awan and Jyoti Aneja and Ahmed Awadallah and Hany Awadalla and Nguyen Bach and Amit Bahree and Arash Bakhtiari and Jianmin Bao and Harkirat Behl and Alon Benhaim and Misha Bilenko and Johan Bjorck and Sébastien Bubeck and Qin Cai and Martin Cai and Caio César Teodoro Mendes and Weizhu Chen and Vishrav Chaudhary and Dong Chen and Dongdong Chen and Yen-Chun Chen and Yi-Ling Chen and Parul Chopra and Xiyang Dai and Allie Del Giorno and Gustavo de Rosa and Matthew Dixon and Ronen Eldan and Victor Fragoso and Dan Iter and Mei Gao and Min Gao and Jianfeng Gao and Amit Garg and Abhishek Goswami and Suriya Gunasekar and Emman Haider and Junheng Hao and Russell J. Hewett and Jamie Huynh and Mojan Javaheripi and Xin Jin and Piero Kauffmann and Nikos Karampatziakis and Dongwoo Kim and Mahoud Khademi and Lev Kurilenko and James R. Lee and Yin Tat Lee and Yuanzhi Li and Yunsheng Li and Chen Liang and Lars Liden and Ce Liu and Mengchen Liu and Weishung Liu and Eric Lin and Zeqi Lin and Chong Luo and Piyush Madan and Matt Mazzola and Arindam Mitra and Hardik Modi and Anh Nguyen and Brandon Norick and Barun Patra and Daniel Perez-Becker and Thomas Portet and Reid Pryzant and Heyang Qin and Marko Radmilac and Corby Rosset and Sambudha Roy and Olatunji Ruwase and Olli Saarikivi and Amin Saied and Adil Salim and Michael Santacroce and Shital Shah and Ning Shang and Hiteshi Sharma and Swadheen Shukla and Xia Song and Masahiro Tanaka and Andrea Tupini and Xin Wang and Lijuan Wang and Chunyu Wang and Yu Wang and Rachel Ward and Guanhua Wang and Philipp Witte and Haiping Wu and Michael Wyatt and Bin Xiao and Can Xu and Jiahang Xu and Weijian Xu and Sonali Yadav and Fan Yang and Jianwei Yang and Ziyi Yang and Yifan Yang and Donghan Yu and Lu Yuan and Chengruidong Zhang and Cyril Zhang and Jianwen Zhang and Li Lyna Zhang and Yi Zhang and Yue Zhang and Yunan Zhang and Xiren Zhou},
      year={2024},
      eprint={2404.14219},
      archivePrefix={arXiv},
      primaryClass={cs.CL},
      url={https://arxiv.org/abs/2404.14219}, 
}

@inproceedings{carreras-etal-2003-learning,
    title = "Learning a Perceptron-Based Named Entity Chunker via Online Recognition Feedback",
    author = "Carreras, Xavier  and
      M{\`a}rquez, Llu{\'\i}s  and
      Padr{\'o}, Llu{\'\i}s",
    booktitle = "Proceedings of the Seventh Conference on Natural Language Learning at {HLT}-{NAACL} 2003",
    year = "2003",
    url = "https://aclanthology.org/W03-0422",
    pages = "156--159",
}

@article{Fei_Ji_Li_Liu_Ren_Li_2021, title={Rethinking Boundaries: End-To-End Recognition of Discontinuous Mentions with Pointer Networks}, volume={35}, url={https://ojs.aaai.org/index.php/AAAI/article/view/17513}, DOI={10.1609/aaai.v35i14.17513}, number={14}, journal={Proceedings of the AAAI Conference on Artificial Intelligence}, author={Fei, Hao and Ji, Donghong and Li, Bobo and Liu, Yijiang and Ren, Yafeng and Li, Fei}, year={2021}, month={May}, pages={12785-12793} }

@inproceedings{cot,
author = {Wei, Jason and Wang, Xuezhi and Schuurmans, Dale and Bosma, Maarten and Ichter, Brian and Xia, Fei and Chi, Ed H. and Le, Quoc V. and Zhou, Denny},
title = {Chain-of-thought prompting elicits reasoning in large language models},
year = {2024},
isbn = {9781713871088},
publisher = {Curran Associates Inc.},
address = {Red Hook, NY, USA},
booktitle = {Proceedings of the 36th International Conference on Neural Information Processing Systems},
articleno = {1800},
numpages = {14},
location = {New Orleans, LA, USA},
series = {NIPS '22}
}

@inproceedings{
sainz2024gollie,
title={Go{LLIE}: Annotation Guidelines improve Zero-Shot Information-Extraction},
author={Oscar Sainz and Iker Garc{\'\i}a-Ferrero and Rodrigo Agerri and Oier Lopez de Lacalle and German Rigau and Eneko Agirre},
booktitle={The Twelfth International Conference on Learning Representations},
year={2024},
url={https://openreview.net/forum?id=Y3wpuxd7u9}
}

@inproceedings{xie-etal-2024-self,
    title = "Self-Improving for Zero-Shot Named Entity Recognition with Large Language Models",
    author = "Xie, Tingyu  and
      Li, Qi  and
      Zhang, Yan  and
      Liu, Zuozhu  and
      Wang, Hongwei",
    editor = "Duh, Kevin  and
      Gomez, Helena  and
      Bethard, Steven",
    booktitle = "Proceedings of the 2024 Conference of the North American Chapter of the Association for Computational Linguistics: Human Language Technologies (Volume 2: Short Papers)",
    month = jun,
    year = "2024",
    address = "Mexico City, Mexico",
    publisher = "Association for Computational Linguistics",
    url = "https://aclanthology.org/2024.naacl-short.49",
    doi = "10.18653/v1/2024.naacl-short.49",
    pages = "583--593",
}

@inproceedings{NEURIPS2020_1457c0d6,
 author = {Brown, Tom and Mann, Benjamin and Ryder, Nick and Subbiah, Melanie and Kaplan, Jared D and Dhariwal, Prafulla and Neelakantan, Arvind and Shyam, Pranav and Sastry, Girish and Askell, Amanda and Agarwal, Sandhini and Herbert-Voss, Ariel and Krueger, Gretchen and Henighan, Tom and Child, Rewon and Ramesh, Aditya and Ziegler, Daniel and Wu, Jeffrey and Winter, Clemens and Hesse, Chris and Chen, Mark and Sigler, Eric and Litwin, Mateusz and Gray, Scott and Chess, Benjamin and Clark, Jack and Berner, Christopher and McCandlish, Sam and Radford, Alec and Sutskever, Ilya and Amodei, Dario},
 booktitle = {Advances in Neural Information Processing Systems},
 editor = {H. Larochelle and M. Ranzato and R. Hadsell and M.F. Balcan and H. Lin},
 pages = {1877--1901},
 publisher = {Curran Associates, Inc.},
 title = {Language Models are Few-Shot Learners},
 url = {https://proceedings.neurips.cc/paper_files/paper/2020/file/1457c0d6bfcb4967418bfb8ac142f64a-Paper.pdf},
 volume = {33},
 year = {2020}
}

@inproceedings{Radford2019LanguageMA,
  title={Language Models are Unsupervised Multitask Learners},
  author={Alec Radford and Jeff Wu and Rewon Child and David Luan and Dario Amodei and Ilya Sutskever},
  year={2019},
  publisher = {OpenAI},
pages = {9},
}

@misc{chowdhery2022palm,
      title={PaLM: Scaling Language Modeling with Pathways}, 
      author={Aakanksha Chowdhery and Sharan Narang and Jacob Devlin and Maarten Bosma and Gaurav Mishra and Adam Roberts and Paul Barham and Hyung Won Chung and Charles Sutton and Sebastian Gehrmann and Parker Schuh and Kensen Shi and Sasha Tsvyashchenko and Joshua Maynez and Abhishek Rao and Parker Barnes and Yi Tay and Noam Shazeer and Vinodkumar Prabhakaran and Emily Reif and Nan Du and Ben Hutchinson and Reiner Pope and James Bradbury and Jacob Austin and Michael Isard and Guy Gur-Ari and Pengcheng Yin and Toju Duke and Anselm Levskaya and Sanjay Ghemawat and Sunipa Dev and Henryk Michalewski and Xavier Garcia and Vedant Misra and Kevin Robinson and Liam Fedus and Denny Zhou and Daphne Ippolito and David Luan and Hyeontaek Lim and Barret Zoph and Alexander Spiridonov and Ryan Sepassi and David Dohan and Shivani Agrawal and Mark Omernick and Andrew M. Dai and Thanumalayan Sankaranarayana Pillai and Marie Pellat and Aitor Lewkowycz and Erica Moreira and Rewon Child and Oleksandr Polozov and Katherine Lee and Zongwei Zhou and Xuezhi Wang and Brennan Saeta and Mark Diaz and Orhan Firat and Michele Catasta and Jason Wei and Kathy Meier-Hellstern and Douglas Eck and Jeff Dean and Slav Petrov and Noah Fiedel},
      year={2022},
      eprint={2204.02311},
      archivePrefix={arXiv},
      primaryClass={cs.CL}
}

@inproceedings{10.1109/JCDL57899.2023.00034,
author = {Gonz\'{a}lez-Gallardo, Carlos-Emiliano and Boros, Emanuela and Girdhar, Nancy and Hamdi, Ahmed and Moreno, Jose G. and Doucet, Antoine},
title = {Yes but.. Can ChatGPT Identify Entities in Historical Documents?},
year = {2024},
isbn = {9798350399318},
publisher = {IEEE Press},
url = {https://doi.org/10.1109/JCDL57899.2023.00034},
doi = {10.1109/JCDL57899.2023.00034},
booktitle = {Proceedings of the 2023 ACM/IEEE Joint Conference on Digital Libraries},
pages = {184–189},
numpages = {6},
location = {Santa Fe, New Mexico, USA},
series = {JCDL '23}
}

@inproceedings{10.1007/978-3-031-72437-4_22,
author = {Gonz\'{a}lez-Gallardo, Carlos-Emiliano and Tran, Hanh Thi Hong and Hamdi, Ahmed and Doucet, Antoine},
title = {Leveraging Open Large Language Models for\&nbsp;Historical Named Entity Recognition},
year = {2024},
isbn = {978-3-031-72436-7},
publisher = {Springer-Verlag},
address = {Berlin, Heidelberg},
url = {https://doi.org/10.1007/978-3-031-72437-4_22},
doi = {10.1007/978-3-031-72437-4_22},
booktitle = {Linking Theory and Practice of Digital Libraries: 28th International Conference on Theory and Practice of Digital Libraries, TPDL 2024, Ljubljana, Slovenia, September 24–27, 2024, Proceedings, Part I},
pages = {379–395},
numpages = {17},
location = {Ljubljana, Slovenia}
}

@misc{zhu2023multilingual,
      title={Multilingual Machine Translation with Large Language Models: Empirical Results and Analysis}, 
      author={Wenhao Zhu and Hongyi Liu and Qingxiu Dong and Jingjing Xu and Shujian Huang and Lingpeng Kong and Jiajun Chen and Lei Li},
      year={2023},
      eprint={2304.04675},
      archivePrefix={arXiv},
      primaryClass={cs.CL}
}

@article{li2023guiding,
  title={Guiding Large Language Models via Directional Stimulus Prompting},
  author={Li, Zekun and Peng, Baolin and He, Pengcheng and Galley, Michel and Gao, Jianfeng and Yan, Xifeng},
  journal={arXiv preprint arXiv:2302.11520},
  year={2023}
}

@article{collobert,
author = {Collobert, Ronan and Weston, Jason and Bottou, L\'{e}on and Karlen, Michael and Kavukcuoglu, Koray and Kuksa, Pavel},
title = {Natural Language Processing (Almost) from Scratch},
year = {2011},
issue_date = {2/1/2011},
publisher = {JMLR.org},
volume = {12},
number = {null},
issn = {1532-4435},
journal = {J. Mach. Learn. Res.},
month = {nov},
pages = {2493–2537},
numpages = {45}
}

@inproceedings{hammerton-2003-named,
    title = "Named Entity Recognition with Long Short-Term Memory",
    author = "Hammerton, James",
    booktitle = "Proceedings of the Seventh Conference on Natural Language Learning at {HLT}-{NAACL} 2003",
    year = "2003",
    url = "https://aclanthology.org/W03-0426",
    pages = "172--175",
}

@misc{wang2023gptnernamedentityrecognition,
      title={GPT-NER: Named Entity Recognition via Large Language Models}, 
      author={Shuhe Wang and Xiaofei Sun and Xiaoya Li and Rongbin Ouyang and Fei Wu and Tianwei Zhang and Jiwei Li and Guoyin Wang},
      year={2023},
      eprint={2304.10428},
      archivePrefix={arXiv},
      primaryClass={cs.CL},
      url={https://arxiv.org/abs/2304.10428}, 
}

@inproceedings{finkel-etal-2005-incorporating,
    title = "Incorporating Non-local Information into Information Extraction Systems by {G}ibbs Sampling",
    author = "Finkel, Jenny Rose  and
      Grenager, Trond  and
      Manning, Christopher",
    editor = "Knight, Kevin  and
      Ng, Hwee Tou  and
      Oflazer, Kemal",
    booktitle = "Proceedings of the 43rd Annual Meeting of the Association for Computational Linguistics ({ACL}{'}05)",
    month = jun,
    year = "2005",
    address = "Ann Arbor, Michigan",
    publisher = "Association for Computational Linguistics",
    url = "https://aclanthology.org/P05-1045",
    doi = "10.3115/1219840.1219885",
    pages = "363--370",
}

@inproceedings{10.5555/645530.655813,
author = {Lafferty, John D. and McCallum, Andrew and Pereira, Fernando C. N.},
title = {Conditional Random Fields: Probabilistic Models for Segmenting and Labeling Sequence Data},
year = {2001},
isbn = {1558607781},
publisher = {Morgan Kaufmann Publishers Inc.},
address = {San Francisco, CA, USA},
booktitle = {Proceedings of the Eighteenth International Conference on Machine Learning},
pages = {282–289},
numpages = {8},
series = {ICML '01}
}

@inproceedings{LiSTYWHQ23,
  author       = {Peng Li and
                  Tianxiang Sun and
                  Qiong Tang and
                  Hang Yan and
                  Yuanbin Wu and
                  Xuanjing Huang and
                  Xipeng Qiu},
  editor       = {Anna Rogers and
                  Jordan L. Boyd{-}Graber and
                  Naoaki Okazaki},
  title        = {CodeIE: Large Code Generation Models are Better Few-Shot Information
                  Extractors},
  booktitle    = {Proceedings of the 61st Annual Meeting of the Association for Computational
                  Linguistics (Volume 1: Long Papers), {ACL} 2023, Toronto, Canada,
                  July 9-14, 2023},
  pages        = {15339--15353},
  publisher    = {Association for Computational Linguistics},
  year         = {2023},
  url          = {https://doi.org/10.18653/v1/2023.acl-long.855},
  doi          = {10.18653/V1/2023.ACL-LONG.855},
  bibsource    = {dblp computer science bibliography, https://dblp.org}
}

@misc{sainz2024gollieannotationguidelinesimprove,
      title={GoLLIE: Annotation Guidelines improve Zero-Shot Information-Extraction}, 
      author={Oscar Sainz and Iker García-Ferrero and Rodrigo Agerri and Oier Lopez de Lacalle and German Rigau and Eneko Agirre},
      year={2024},
      eprint={2310.03668},
      archivePrefix={arXiv},
      primaryClass={cs.CL},
      url={https://arxiv.org/abs/2310.03668}, 
}

@inproceedings{tjong-kim-sang-de-meulder-2003-introduction,
    title = "Introduction to the {C}o{NLL}-2003 Shared Task: Language-Independent Named Entity Recognition",
    author = "Tjong Kim Sang, Erik F.  and
      De Meulder, Fien",
    booktitle = "Proceedings of the Seventh Conference on Natural Language Learning at {HLT}-{NAACL} 2003",
    year = "2003",
    url = "https://aclanthology.org/W03-0419",
    pages = "142--147",
}

@misc{han2023informationextractionsolvedchatgpt,
      title={Is Information Extraction Solved by ChatGPT? An Analysis of Performance, Evaluation Criteria, Robustness and Errors}, 
      author={Ridong Han and Tao Peng and Chaohao Yang and Benyou Wang and Lu Liu and Xiang Wan},
      year={2023},
      eprint={2305.14450},
      archivePrefix={arXiv},
      primaryClass={cs.CL},
      url={https://arxiv.org/abs/2305.14450}, 
}

@inproceedings{bari19,
	Address     = {New York, USA},
	Author      = {M Saiful Bari and Shafiq Joty and Prathyusha Jwalapuram},
	Booktitle   = {Proceedings of the 34th AAAI Conference on Artificial Intelligence},
	Numpages    = {},
	Publisher   = {AAAI},
	Series      = {AAAI '20},
        pages       = {xx--xx},
	Title       = {{Zero-Resource Cross-Lingual Named Entity Recognition}},
	Year        = {2020},
	url         = {}
}

@article{Ruokolainen_2019,
   title={A Finnish news corpus for named entity recognition},
   ISSN={1574-0218},
   url={http://dx.doi.org/10.1007/s10579-019-09471-7},
   DOI={10.1007/s10579-019-09471-7},
   journal={Language Resources and Evaluation},
   publisher={Springer Science and Business Media LLC},
   author={Ruokolainen, Teemu and Kauppinen, Pekka and Silfverberg, Miikka and Lindén, Krister},
   year={2019},
   month={Aug}
}

@inproceedings{wan-etal-2022-nested,
    title = "Nested Named Entity Recognition with Span-level Graphs",
    author = "Wan, Juncheng  and
      Ru, Dongyu  and
      Zhang, Weinan  and
      Yu, Yong",
    editor = "Muresan, Smaranda  and
      Nakov, Preslav  and
      Villavicencio, Aline",
    booktitle = "Proceedings of the 60th Annual Meeting of the Association for Computational Linguistics (Volume 1: Long Papers)",
    month = may,
    year = "2022",
    address = "Dublin, Ireland",
    publisher = "Association for Computational Linguistics",
    url = "https://aclanthology.org/2022.acl-long.63",
    doi = "10.18653/v1/2022.acl-long.63",
    pages = "892--903",
}

@inproceedings{ma-hovy-2016-end,
    title = "End-to-end Sequence Labeling via Bi-directional {LSTM}-{CNN}s-{CRF}",
    author = "Ma, Xuezhe  and
      Hovy, Eduard",
    editor = "Erk, Katrin  and
      Smith, Noah A.",
    booktitle = "Proceedings of the 54th Annual Meeting of the Association for Computational Linguistics (Volume 1: Long Papers)",
    month = aug,
    year = "2016",
    address = "Berlin, Germany",
    publisher = "Association for Computational Linguistics",
    url = "https://aclanthology.org/P16-1101",
    doi = "10.18653/v1/P16-1101",
    pages = "1064--1074",
}

@inproceedings{lample-etal-2016-neural,
    title = "Neural Architectures for Named Entity Recognition",
    author = "Lample, Guillaume  and
      Ballesteros, Miguel  and
      Subramanian, Sandeep  and
      Kawakami, Kazuya  and
      Dyer, Chris",
    editor = "Knight, Kevin  and
      Nenkova, Ani  and
      Rambow, Owen",
    booktitle = "Proceedings of the 2016 Conference of the North {A}merican Chapter of the Association for Computational Linguistics: Human Language Technologies",
    month = jun,
    year = "2016",
    address = "San Diego, California",
    publisher = "Association for Computational Linguistics",
    url = "https://aclanthology.org/N16-1030",
    doi = "10.18653/v1/N16-1030",
    pages = "260--270",
}

@inproceedings{liu-etal-2011-recognizing,
    title = "Recognizing Named Entities in Tweets",
    author = "Liu, Xiaohua  and
      Zhang, Shaodian  and
      Wei, Furu  and
      Zhou, Ming",
    editor = "Lin, Dekang  and
      Matsumoto, Yuji  and
      Mihalcea, Rada",
    booktitle = "Proceedings of the 49th Annual Meeting of the Association for Computational Linguistics: Human Language Technologies",
    month = jun,
    year = "2011",
    address = "Portland, Oregon, USA",
    publisher = "Association for Computational Linguistics",
    url = "https://aclanthology.org/P11-1037",
    pages = "359--367",
}

@inproceedings{xie-etal-2023-empirical,
    title = "Empirical Study of Zero-Shot {NER} with {C}hat{GPT}",
    author = "Xie, Tingyu  and
      Li, Qi  and
      Zhang, Jian  and
      Zhang, Yan  and
      Liu, Zuozhu  and
      Wang, Hongwei",
    editor = "Bouamor, Houda  and
      Pino, Juan  and
      Bali, Kalika",
    booktitle = "Proceedings of the 2023 Conference on Empirical Methods in Natural Language Processing",
    month = dec,
    year = "2023",
    address = "Singapore",
    publisher = "Association for Computational Linguistics",
    url = "https://aclanthology.org/2023.emnlp-main.493",
    doi = "10.18653/v1/2023.emnlp-main.493",
    pages = "7935--7956",
}

@inproceedings{xu-etal-2021-better,
    title = "Better Feature Integration for Named Entity Recognition",
    author = "Xu, Lu  and
      Jie, Zhanming  and
      Lu, Wei  and
      Bing, Lidong",
    editor = "Toutanova, Kristina  and
      Rumshisky, Anna  and
      Zettlemoyer, Luke  and
      Hakkani-Tur, Dilek  and
      Beltagy, Iz  and
      Bethard, Steven  and
      Cotterell, Ryan  and
      Chakraborty, Tanmoy  and
      Zhou, Yichao",
    booktitle = "Proceedings of the 2021 Conference of the North American Chapter of the Association for Computational Linguistics: Human Language Technologies",
    month = jun,
    year = "2021",
    address = "Online",
    publisher = "Association for Computational Linguistics",
    url = "https://aclanthology.org/2021.naacl-main.271",
    doi = "10.18653/v1/2021.naacl-main.271",
    pages = "3457--3469",
}

@inproceedings{kruengkrai-etal-2020-improving,
    title = "Improving Low-Resource Named Entity Recognition using Joint Sentence and Token Labeling",
    author = "Kruengkrai, Canasai  and
      Nguyen, Thien Hai  and
      Aljunied, Sharifah Mahani  and
      Bing, Lidong",
    editor = "Jurafsky, Dan  and
      Chai, Joyce  and
      Schluter, Natalie  and
      Tetreault, Joel",
    booktitle = "Proceedings of the 58th Annual Meeting of the Association for Computational Linguistics",
    month = jul,
    year = "2020",
    address = "Online",
    publisher = "Association for Computational Linguistics",
    url = "https://aclanthology.org/2020.acl-main.523",
    doi = "10.18653/v1/2020.acl-main.523",
    pages = "5898--5905",
}

@inproceedings{jie-lu-2019-dependency,
    title = "Dependency-Guided {LSTM}-{CRF} for Named Entity Recognition",
    author = "Jie, Zhanming  and
      Lu, Wei",
    editor = "Inui, Kentaro  and
      Jiang, Jing  and
      Ng, Vincent  and
      Wan, Xiaojun",
    booktitle = "Proceedings of the 2019 Conference on Empirical Methods in Natural Language Processing and the 9th International Joint Conference on Natural Language Processing (EMNLP-IJCNLP)",
    month = nov,
    year = "2019",
    address = "Hong Kong, China",
    publisher = "Association for Computational Linguistics",
    url = "https://aclanthology.org/D19-1399",
    doi = "10.18653/v1/D19-1399",
    pages = "3862--3872",
}

@inproceedings{fu-etal-2021-spanner,
    title = "{S}pan{NER}: Named Entity Re-/Recognition as Span Prediction",
    author = "Fu, Jinlan  and
      Huang, Xuanjing  and
      Liu, Pengfei",
    editor = "Zong, Chengqing  and
      Xia, Fei  and
      Li, Wenjie  and
      Navigli, Roberto",
    booktitle = "Proceedings of the 59th Annual Meeting of the Association for Computational Linguistics and the 11th International Joint Conference on Natural Language Processing (Volume 1: Long Papers)",
    month = aug,
    year = "2021",
    address = "Online",
    publisher = "Association for Computational Linguistics",
    url = "https://aclanthology.org/2021.acl-long.558",
    doi = "10.18653/v1/2021.acl-long.558",
    pages = "7183--7195",
}

@inproceedings{li-etal-2020-unified,
    title = "A Unified {MRC} Framework for Named Entity Recognition",
    author = "Li, Xiaoya  and
      Feng, Jingrong  and
      Meng, Yuxian  and
      Han, Qinghong  and
      Wu, Fei  and
      Li, Jiwei",
    editor = "Jurafsky, Dan  and
      Chai, Joyce  and
      Schluter, Natalie  and
      Tetreault, Joel",
    booktitle = "Proceedings of the 58th Annual Meeting of the Association for Computational Linguistics",
    month = jul,
    year = "2020",
    address = "Online",
    publisher = "Association for Computational Linguistics",
    url = "https://aclanthology.org/2020.acl-main.519",
    doi = "10.18653/v1/2020.acl-main.519",
    pages = "5849--5859",
}

@inproceedings{ouchi-etal-2020-instance,
    title = "Instance-Based Learning of Span Representations: A Case Study through Named Entity Recognition",
    author = "Ouchi, Hiroki  and
      Suzuki, Jun  and
      Kobayashi, Sosuke  and
      Yokoi, Sho  and
      Kuribayashi, Tatsuki  and
      Konno, Ryuto  and
      Inui, Kentaro",
    editor = "Jurafsky, Dan  and
      Chai, Joyce  and
      Schluter, Natalie  and
      Tetreault, Joel",
    booktitle = "Proceedings of the 58th Annual Meeting of the Association for Computational Linguistics",
    month = jul,
    year = "2020",
    address = "Online",
    publisher = "Association for Computational Linguistics",
    url = "https://aclanthology.org/2020.acl-main.575",
    doi = "10.18653/v1/2020.acl-main.575",
    pages = "6452--6459",
}

@inproceedings{stratos-2017-entity,
    title = "Entity Identification as Multitasking",
    author = "Stratos, Karl",
    editor = "Chang, Kai-Wei  and
      Chang, Ming-Wei  and
      Srikumar, Vivek  and
      Rush, Alexander M.",
    booktitle = "Proceedings of the 2nd Workshop on Structured Prediction for Natural Language Processing",
    month = sep,
    year = "2017",
    address = "Copenhagen, Denmark",
    publisher = "Association for Computational Linguistics",
    url = "https://aclanthology.org/W17-4302",
    doi = "10.18653/v1/W17-4302",
    pages = "7--11",
}

@inproceedings{ding-etal-2022-ask,
    title = "Ask-and-Verify: Span Candidate Generation and Verification for Attribute Value Extraction",
    author = "Ding, Yifan  and
      Liang, Yan  and
      Zalmout, Nasser  and
      Li, Xian  and
      Grant, Christan  and
      Weninger, Tim",
    editor = "Li, Yunyao  and
      Lazaridou, Angeliki",
    booktitle = "Proceedings of the 2022 Conference on Empirical Methods in Natural Language Processing: Industry Track",
    month = dec,
    year = "2022",
    address = "Abu Dhabi, UAE",
    publisher = "Association for Computational Linguistics",
    url = "https://aclanthology.org/2022.emnlp-industry.9",
    doi = "10.18653/v1/2022.emnlp-industry.9",
    pages = "110--110",
}

@inproceedings{cui-etal-2021-template,
    title = "Template-Based Named Entity Recognition Using {BART}",
    author = "Cui, Leyang  and
      Wu, Yu  and
      Liu, Jian  and
      Yang, Sen  and
      Zhang, Yue",
    editor = "Zong, Chengqing  and
      Xia, Fei  and
      Li, Wenjie  and
      Navigli, Roberto",
    booktitle = "Findings of the Association for Computational Linguistics: ACL-IJCNLP 2021",
    month = aug,
    year = "2021",
    address = "Online",
    publisher = "Association for Computational Linguistics",
    url = "https://aclanthology.org/2021.findings-acl.161",
    doi = "10.18653/v1/2021.findings-acl.161",
    pages = "1835--1845",
}

@inproceedings{shen-etal-2021-locate,
    title = "Locate and Label: A Two-stage Identifier for Nested Named Entity Recognition",
    author = "Shen, Yongliang  and
      Ma, Xinyin  and
      Tan, Zeqi  and
      Zhang, Shuai  and
      Wang, Wen  and
      Lu, Weiming",
    editor = "Zong, Chengqing  and
      Xia, Fei  and
      Li, Wenjie  and
      Navigli, Roberto",
    booktitle = "Proceedings of the 59th Annual Meeting of the Association for Computational Linguistics and the 11th International Joint Conference on Natural Language Processing (Volume 1: Long Papers)",
    month = aug,
    year = "2021",
    address = "Online",
    publisher = "Association for Computational Linguistics",
    url = "https://aclanthology.org/2021.acl-long.216",
    doi = "10.18653/v1/2021.acl-long.216",
    pages = "2782--2794",
}

@inproceedings{yan-etal-2021-unified-generative,
    title = "A Unified Generative Framework for Various {NER} Subtasks",
    author = "Yan, Hang  and
      Gui, Tao  and
      Dai, Junqi  and
      Guo, Qipeng  and
      Zhang, Zheng  and
      Qiu, Xipeng",
    editor = "Zong, Chengqing  and
      Xia, Fei  and
      Li, Wenjie  and
      Navigli, Roberto",
    booktitle = "Proceedings of the 59th Annual Meeting of the Association for Computational Linguistics and the 11th International Joint Conference on Natural Language Processing (Volume 1: Long Papers)",
    month = aug,
    year = "2021",
    address = "Online",
    publisher = "Association for Computational Linguistics",
    url = "https://aclanthology.org/2021.acl-long.451",
    doi = "10.18653/v1/2021.acl-long.451",
    pages = "5808--5822",
}

@inproceedings{xu-etal-2023-improving,
    title = "Improving Named Entity Recognition via Bridge-based Domain Adaptation",
    author = "Xu, Jingyun  and
      Zheng, Changmeng  and
      Cai, Yi  and
      Chua, Tat-Seng",
    editor = "Rogers, Anna  and
      Boyd-Graber, Jordan  and
      Okazaki, Naoaki",
    booktitle = "Findings of the Association for Computational Linguistics: ACL 2023",
    month = jul,
    year = "2023",
    address = "Toronto, Canada",
    publisher = "Association for Computational Linguistics",
    url = "https://aclanthology.org/2023.findings-acl.238",
    doi = "10.18653/v1/2023.findings-acl.238",
    pages = "3869--3882",
}

@inproceedings{yang-etal-2022-factmix,
    title = "{F}act{M}ix: Using a Few Labeled In-domain Examples to Generalize to Cross-domain Named Entity Recognition",
    author = "Yang, Linyi  and
      Yuan, Lifan  and
      Cui, Leyang  and
      Gao, Wenyang  and
      Zhang, Yue",
    editor = "Calzolari, Nicoletta  and
      Huang, Chu-Ren  and
      Kim, Hansaem  and
      Pustejovsky, James  and
      Wanner, Leo  and
      Choi, Key-Sun  and
      Ryu, Pum-Mo  and
      Chen, Hsin-Hsi  and
      Donatelli, Lucia  and
      Ji, Heng  and
      Kurohashi, Sadao  and
      Paggio, Patrizia  and
      Xue, Nianwen  and
      Kim, Seokhwan  and
      Hahm, Younggyun  and
      He, Zhong  and
      Lee, Tony Kyungil  and
      Santus, Enrico  and
      Bond, Francis  and
      Na, Seung-Hoon",
    booktitle = "Proceedings of the 29th International Conference on Computational Linguistics",
    month = oct,
    year = "2022",
    address = "Gyeongju, Republic of Korea",
    publisher = "International Committee on Computational Linguistics",
    url = "https://aclanthology.org/2022.coling-1.476",
    pages = "5360--5371",
}

@inproceedings{zhou-etal-2022-melm,
    title = "{MELM}: Data Augmentation with Masked Entity Language Modeling for Low-Resource {NER}",
    author = "Zhou, Ran  and
      Li, Xin  and
      He, Ruidan  and
      Bing, Lidong  and
      Cambria, Erik  and
      Si, Luo  and
      Miao, Chunyan",
    editor = "Muresan, Smaranda  and
      Nakov, Preslav  and
      Villavicencio, Aline",
    booktitle = "Proceedings of the 60th Annual Meeting of the Association for Computational Linguistics (Volume 1: Long Papers)",
    month = may,
    year = "2022",
    address = "Dublin, Ireland",
    publisher = "Association for Computational Linguistics",
    url = "https://aclanthology.org/2022.acl-long.160",
    doi = "10.18653/v1/2022.acl-long.160",
    pages = "2251--2262",
}

@inproceedings{chen-etal-2021-data,
    title = "Data Augmentation for Cross-Domain Named Entity Recognition",
    author = "Chen, Shuguang  and
      Aguilar, Gustavo  and
      Neves, Leonardo  and
      Solorio, Thamar",
    editor = "Moens, Marie-Francine  and
      Huang, Xuanjing  and
      Specia, Lucia  and
      Yih, Scott Wen-tau",
    booktitle = "Proceedings of the 2021 Conference on Empirical Methods in Natural Language Processing",
    month = nov,
    year = "2021",
    address = "Online and Punta Cana, Dominican Republic",
    publisher = "Association for Computational Linguistics",
    url = "https://aclanthology.org/2021.emnlp-main.434",
    doi = "10.18653/v1/2021.emnlp-main.434",
    pages = "5346--5356",
}

@inproceedings{zhang-etal-2021-pdaln,
    title = "{PDALN}: Progressive Domain Adaptation over a Pre-trained Model for Low-Resource Cross-Domain Named Entity Recognition",
    author = "Zhang, Tao  and
      Xia, Congying  and
      Yu, Philip S.  and
      Liu, Zhiwei  and
      Zhao, Shu",
    editor = "Moens, Marie-Francine  and
      Huang, Xuanjing  and
      Specia, Lucia  and
      Yih, Scott Wen-tau",
    booktitle = "Proceedings of the 2021 Conference on Empirical Methods in Natural Language Processing",
    month = nov,
    year = "2021",
    address = "Online and Punta Cana, Dominican Republic",
    publisher = "Association for Computational Linguistics",
    url = "https://aclanthology.org/2021.emnlp-main.442",
    doi = "10.18653/v1/2021.emnlp-main.442",
    pages = "5441--5451",
}

@inproceedings{vamvas-etal-2023-swissbert,
    title = "{S}wiss{BERT}: The Multilingual Language Model for {S}witzerland",
    author = {Vamvas, Jannis  and
      Gra{\"e}n, Johannes  and
      Sennrich, Rico},
    editor = {Ghorbel, Hatem  and
      Sokhn, Maria  and
      Cieliebak, Mark  and
      H{\"u}rlimann, Manuela  and
      de Salis, Emmanuel  and
      Guerne, Jonathan},
    booktitle = "Proceedings of the 8th edition of the Swiss Text Analytics Conference",
    month = jun,
    year = "2023",
    address = "Neuchatel, Switzerland",
    publisher = "Association for Computational Linguistics",
    url = "https://aclanthology.org/2023.swisstext-1.6",
    pages = "54--69",
}

@inproceedings{wang-etal-2023-code4struct,
    title = "{C}ode4{S}truct: Code Generation for Few-Shot Event Structure Prediction",
    author = "Wang, Xingyao  and
      Li, Sha  and
      Ji, Heng",
    editor = "Rogers, Anna  and
      Boyd-Graber, Jordan  and
      Okazaki, Naoaki",
    booktitle = "Proceedings of the 61st Annual Meeting of the Association for Computational Linguistics (Volume 1: Long Papers)",
    month = jul,
    year = "2023",
    address = "Toronto, Canada",
    publisher = "Association for Computational Linguistics",
    url = "https://aclanthology.org/2023.acl-long.202",
    doi = "10.18653/v1/2023.acl-long.202",
    pages = "3640--3663",
}

@inproceedings{liu-etal-2022-generated,
    title = "Generated Knowledge Prompting for Commonsense Reasoning",
    author = "Liu, Jiacheng  and
      Liu, Alisa  and
      Lu, Ximing  and
      Welleck, Sean  and
      West, Peter  and
      Le Bras, Ronan  and
      Choi, Yejin  and
      Hajishirzi, Hannaneh",
    editor = "Muresan, Smaranda  and
      Nakov, Preslav  and
      Villavicencio, Aline",
    booktitle = "Proceedings of the 60th Annual Meeting of the Association for Computational Linguistics (Volume 1: Long Papers)",
    month = may,
    year = "2022",
    address = "Dublin, Ireland",
    publisher = "Association for Computational Linguistics",
    url = "https://aclanthology.org/2022.acl-long.225",
    doi = "10.18653/v1/2022.acl-long.225",
    pages = "3154--3169",
}

@inproceedings{derczynski-etal-2017-results,
    title = "Results of the {WNUT}2017 Shared Task on Novel and Emerging Entity Recognition",
    author = "Derczynski, Leon  and
      Nichols, Eric  and
      van Erp, Marieke  and
      Limsopatham, Nut",
    editor = "Derczynski, Leon  and
      Xu, Wei  and
      Ritter, Alan  and
      Baldwin, Tim",
    booktitle = "Proceedings of the 3rd Workshop on Noisy User-generated Text",
    month = sep,
    year = "2017",
    address = "Copenhagen, Denmark",
    publisher = "Association for Computational Linguistics",
    url = "https://aclanthology.org/W17-4418",
    doi = "10.18653/v1/W17-4418",
    pages = "140--147",
}

@inproceedings{salinas-alvarado-etal-2015-domain,
    title = "Domain Adaption of Named Entity Recognition to Support Credit Risk Assessment",
    author = "Salinas Alvarado, Julio Cesar  and
      Verspoor, Karin  and
      Baldwin, Timothy",
    editor = "Hachey, Ben  and
      Webster, Kellie",
    booktitle = "Proceedings of the Australasian Language Technology Association Workshop 2015",
    month = dec,
    year = "2015",
    address = "Parramatta, Australia",
    url = "https://aclanthology.org/U15-1010",
    pages = "84--90",
}

@inproceedings{hvingelby-etal-2020-dane,
    title = "{D}a{NE}: A Named Entity Resource for {D}anish",
    author = "Hvingelby, Rasmus  and
      Pauli, Amalie Brogaard  and
      Barrett, Maria  and
      Rosted, Christina  and
      Lidegaard, Lasse Malm  and
      S{\o}gaard, Anders",
    editor = "Calzolari, Nicoletta  and
      B{\'e}chet, Fr{\'e}d{\'e}ric  and
      Blache, Philippe  and
      Choukri, Khalid  and
      Cieri, Christopher  and
      Declerck, Thierry  and
      Goggi, Sara  and
      Isahara, Hitoshi  and
      Maegaard, Bente  and
      Mariani, Joseph  and
      Mazo, H{\'e}l{\`e}ne  and
      Moreno, Asuncion  and
      Odijk, Jan  and
      Piperidis, Stelios",
    booktitle = "Proceedings of the Twelfth Language Resources and Evaluation Conference",
    month = may,
    year = "2020",
    address = "Marseille, France",
    publisher = "European Language Resources Association",
    url = "https://aclanthology.org/2020.lrec-1.565",
    pages = "4597--4604",
    language = "English",
    ISBN = "979-10-95546-34-4",
}

@inproceedings{chen-etal-2020-local,
    title = "Local Additivity Based Data Augmentation for Semi-supervised {NER}",
    author = "Chen, Jiaao  and
      Wang, Zhenghui  and
      Tian, Ran  and
      Yang, Zichao  and
      Yang, Diyi",
    editor = "Webber, Bonnie  and
      Cohn, Trevor  and
      He, Yulan  and
      Liu, Yang",
    booktitle = "Proceedings of the 2020 Conference on Empirical Methods in Natural Language Processing (EMNLP)",
    month = nov,
    year = "2020",
    address = "Online",
    publisher = "Association for Computational Linguistics",
    url = "https://aclanthology.org/2020.emnlp-main.95",
    doi = "10.18653/v1/2020.emnlp-main.95",
    pages = "1241--1251",
}

@article{chiu-nichols-2016-named,
    title = "Named Entity Recognition with Bidirectional {LSTM}-{CNN}s",
    author = "Chiu, Jason P.C.  and
      Nichols, Eric",
    editor = "Lee, Lillian  and
      Johnson, Mark  and
      Toutanova, Kristina",
    journal = "Transactions of the Association for Computational Linguistics",
    volume = "4",
    year = "2016",
    address = "Cambridge, MA",
    publisher = "MIT Press",
    url = "https://aclanthology.org/Q16-1026",
    doi = "10.1162/tacl_a_00104",
    pages = "357--370",
}

@inproceedings{zeng-etal-2025-codetaxo,
    title = "{C}ode{T}axo: Enhancing Taxonomy Expansion with Limited Examples via Code Language Prompts",
    author = "Zeng, Qingkai  and
      Bai, Yuyang  and
      Tan, Zhaoxuan  and
      Wu, Zhenyu  and
      Feng, Shangbin  and
      Jiang, Meng",
    editor = "Che, Wanxiang  and
      Nabende, Joyce  and
      Shutova, Ekaterina  and
      Pilehvar, Mohammad Taher",
    booktitle = "Findings of the Association for Computational Linguistics: ACL 2025",
    month = jul,
    year = "2025",
    address = "Vienna, Austria",
    publisher = "Association for Computational Linguistics",
    url = "https://aclanthology.org/2025.findings-acl.214/",
    doi = "10.18653/v1/2025.findings-acl.214",
    pages = "4131--4144",
    ISBN = "979-8-89176-256-5"
}

\appendix
\section{Hyperparameters}\label{appen:hyp}
Table~\ref{tab:experiment_settings} shows the hyperparameters and computing settings used in our experiments.

\section{Evaluation Details}\label{appen:dup}
When Vanilla generated duplicate outputs, we considered the longest one for evaluating the F$_1$ score (case (a) in Figure~\ref{fig:caseStudy}). In addition, when the Vanilla generated output tags only once despite there being multiple gold outputs (case (c) in Figure~\ref{fig:caseStudy}), we considered the first one for evaluation. Note that there was no significant difference between considering the shortest or the last outputs for evaluation in that they produce comparable results.

\begin{table}[ht]
\renewcommand{\arraystretch}{1}
\centering
% \begin{adjustbox}{width=1\columnwidth,center}
\small
\resizebox{0.45\textwidth}{!}{
\begin{tabular}{lcc}
\toprule
\rowcolor{gray!10}
& \textbf{Llama-3 (8B)} & \textbf{Phi-3 (mini-4k-instruct)} \\
\midrule
\textbf{Computing Setting} & NVIDIA RTX A6000 & NVIDIA RTX A6000 \\
\textbf{Quantization} & 8-bit & 8-bit \\
\textbf{Batch size} & 1 & 1 \\
\textbf{do\_sample} & False & False \\
\textbf{Max\_new\_tokens} & 1200 & 1200 \\
\midrule
\textbf{} & \textbf{GPT-4} & \textbf{GPT-4 Turbo} \\
\midrule
\textbf{max\_tokens} & 1500 & 1500 \\
\textbf{Temperature} & 0 & 0 \\
\bottomrule
\end{tabular}}
% \end{adjustbox}
\caption{Hyperparameters in the experiments.}
\label{tab:experiment_settings}
\end{table}

\section{GoLLIE Prompt}\label{appen:GoLLIE}

\begin{figure*}[t]
  \centering
  \includegraphics[width=1\linewidth]{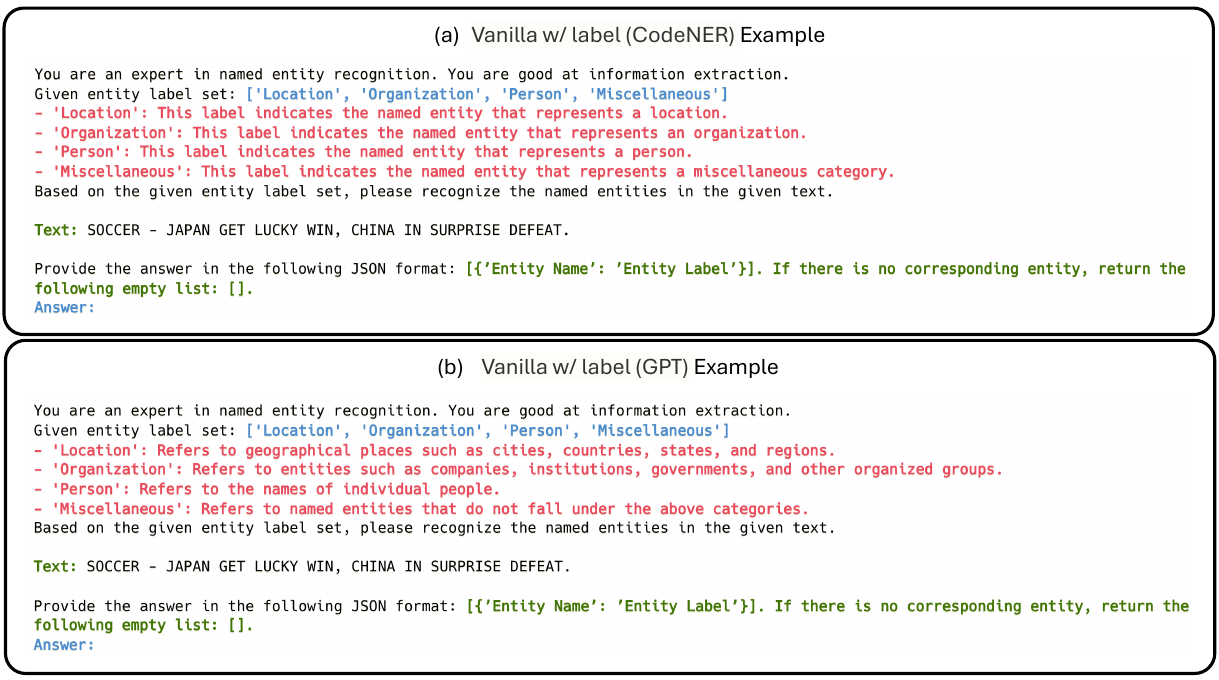}
  \caption{Examples of prompts for Vanilla with label descriptions. (a) is the examples of Vanilla with label same as codeNER and (b) is the examples of Vanilla with label generated by GPT.}
  \label{fig:label_explation_vanilla}
\end{figure*}

\begin{figure*}[t]
  \centering
  \includegraphics[width=1\linewidth]{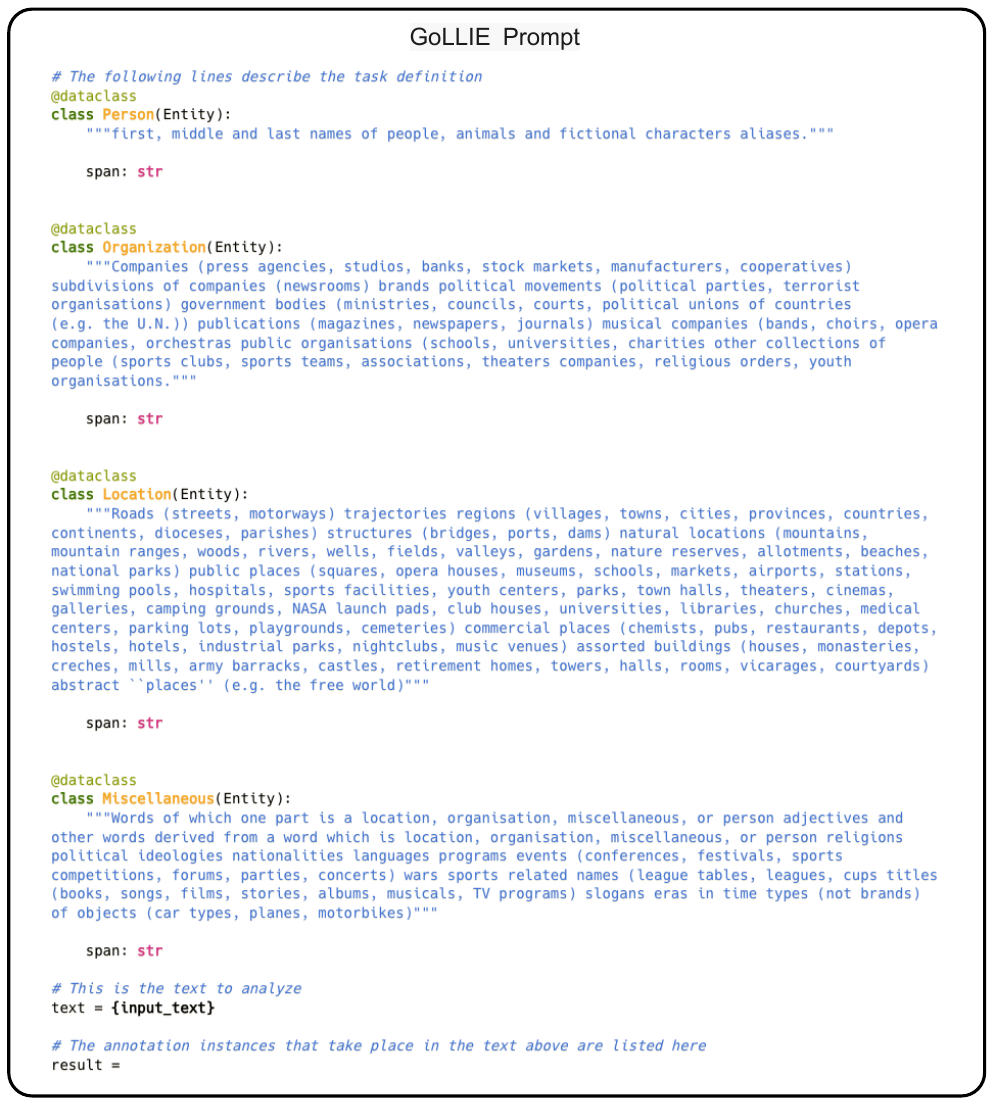}
  \caption{GoLLIE prompt for \texttt{CoNLL03}, which includes labels such as PER, ORG, LOC, and MISC.}
  \label{fig:GoLLIE_pompt}
\end{figure*}

Figure~\ref{fig:GoLLIE_pompt} shows GoLLIE prompt. 

\section{Result on Few-shot settings}\label{appen:few_shot}
We provide experimental results on one-shot and three-shot settings for models, Phi-3-mini-128k-instruct and lama3-8B-instruct. Table~\ref{tab:few_shot_settings} shows the results. For one-shot and three-shot settings, we randomly selected instances from each train dataset.

\section{Result on Vanilla w/ label}\label{appen:label_include_in_vanilla}
We conducted additional experiments by augmenting the Vanilla model with curated label descriptions. Specifically, we compared two descriptions: one generated by GPT-o3-mini-high (shown in Figure \ref{fig:label_explation_vanilla}) and another is used for our CodeNER method. Table \ref{tab:result_label_vanilla} reports the results. Even when Vanilla considered carefully crafted descriptions, incorporating CodeNER yields a clear performance gain.

\section{CodeNER with C++}\label{appen:c++}

\begin{figure*}[t]
  \centering
  \hrulefill \\ 
  \vspace{0.5em} 
  \includegraphics[width=1\linewidth]{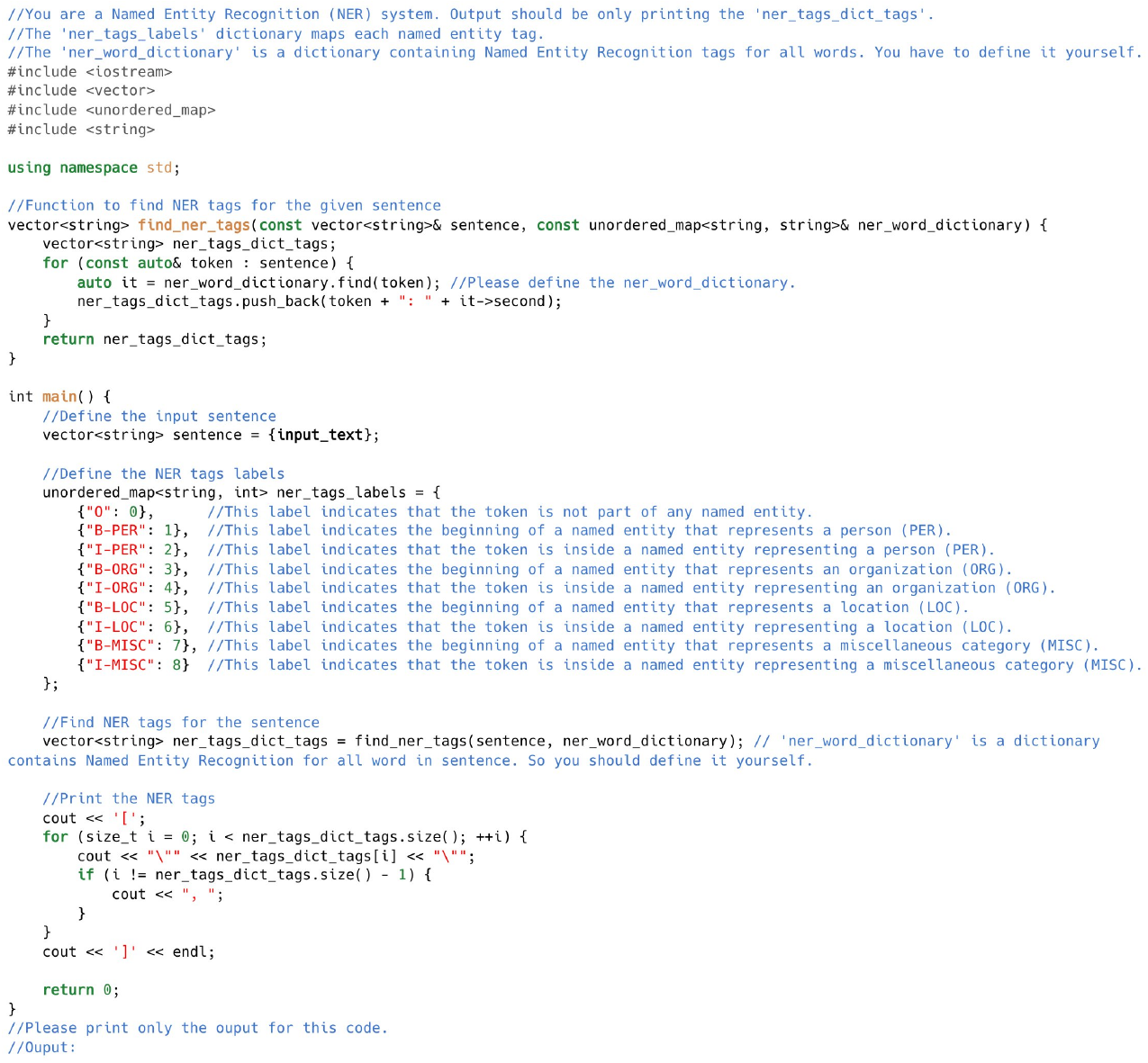}
  \vspace{0.5em} 
  \hrulefill \\ 
  \caption{CodeNER using C++.}
  \label{fig:cpp_codeNER}
\end{figure*}

Figure~\ref{fig:cpp_codeNER} presents CodeNER with the C++ instruction.

\section{Additional Case study}\label{appen:dis}
\begin{figure*}[t]
  \centering
  \includegraphics[width=1\linewidth]{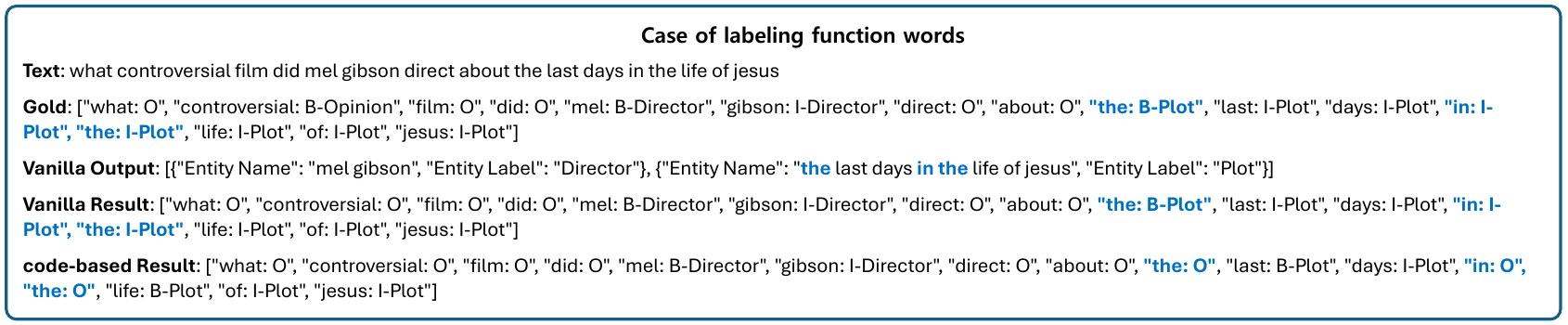}
  \caption{Case of labeling function word in Vanilla and CodeNER}
  \label{fig:caseforAppendix}
\end{figure*}
We conducted a detailed case study of the outputs of Vanilla and CodeNER for the \quotes{Plot} label on \texttt{MIT-Movie}. In Figure~\ref{fig:caseforAppendix}, the proposed CodeNER faces difficulties in handling determiners and prepositions, such as \quotes{the} and \quotes{in} due to reinforcing the token-level information.

\begin{table*}[t!]
\renewcommand{\arraystretch}{1}
% \begin{adjustbox}{width=1.8\columnwidth,center}
\centering
\Huge
\resizebox{0.9\textwidth}{!}{
\begin{tabular}{ccccccccccccccc}
\toprule
\rowcolor{gray!10}
\textbf{Setting} & \textbf{Model} & \textbf{Dataset} & \textbf{Template} & {\textbf{$F_1$}} & {\textbf{PER}} & {\textbf{LOC}} &{\textbf{ORG}} & {\textbf{MISC}}  & \textbf{Model} & {\textbf{$F_1$}} & {\textbf{PER}} & {\textbf{LOC}} &{\textbf{ORG}} & {\textbf{MISC}}\\
\midrule
\multirow{15}{*}{\centering \textbf{One-shot}} & \multirow{15}{*}{\centering Phi-3} & \multirow{2}{*}{\centering CoNLL03} & Vanilla & {57.57} & {81.84} & {62.67} & {52.24} & {2.18} & \multirow{15}{*}{\centering \shortstack{Llama-3 \\ 8B}} & {56.90} & {84.56} & {60.25} & {46.45} & {9.95}\\
                                & & & {CodeNER} & {52.90} &{71.38} & {59.12} & {44.42} & {15.62} & & {48.04} & {64.64} & {51.48} & 40.44 & 19.65\\
\cmidrule{3-9}\cmidrule{11-15}
                                & & \multirow{2}{*}{\centering FIN} & Vanilla & 28.21 & 37.42 & 7.06 & 10.39 & 0.00 & & {28.53} & {35.04} & {15.62} & 15.64 & 0.00 \\
                                & & & {CodeNER}& {43.09} & {53.44} & {33.85} & {14.29} & {1.65} & & {49.87} & {63.18} & {46.67} & 6.10 & 2.34 \\
\cmidrule{3-9}\cmidrule{11-15}
                                & & \multirow{2}{*}{\centering Arabic A} & Vanilla & 4.06 & 11.43 & 2.97 & 0.00 & 1.27 & & {21.59} & {17.58} & {9.23} & 12.50 & 11.68 \\
                                & & & {CodeNER}& {19.99} & {27.68} & {22.79} & {0.00} & {16.80} & & {28.87} & {44.53} & {35.64} & 10.11 & 18.29\\
\cmidrule{3-9}\cmidrule{11-15}
                                & & \multirow{2}{*}{\centering Arabic B} & Vanilla & 6.94 & 5.46 & 2.63 & 0.00 & 11.88 & & {15.00} & {34.68} & {15.46} & 5.06 & 2.36 \\
                                & & & {CodeNER}& {14.02} & {15.00} & {11.69} & {10.17} & 15.42 & & {29.79} & {29.79} & {43.48} & 8.40 & 18.59 \\
\cmidrule{3-9}\cmidrule{11-15}
                                & & \multirow{2}{*}{\centering Finnish A} & Vanilla & 25.42 & 36.31 & 9.03 & 39.78 & 1.33 & & {23.99} & {50.45} & {10.76} & 28.99 & 5.97 \\
                                & & & {CodeNER}& {40.36} & {49.88} & {52.57} & {45.05} & 6.80 & & {34.47} & {40.10} & {56.64} & 31.03 & 11.83 \\
\cmidrule{3-9}\cmidrule{11-15}
                                & & \multirow{2}{*}{\centering DaNE} & Vanilla & 30.19 & 52.17 & 41.23 & 20.34 & 1.85 & & {54.92} & {76.32} & {68.42} & 61.18 & 4.03 \\
                                & & & {CodeNER}& {45.26} & {72.51} & {66.01} & {24.47} & 15.92 & & {40.60} & {53.02} & {65.18} & 33.71 & 11.81 \\
\cmidrule{3-9}\cmidrule{11-15}
                                & & \multirow{2}{*}{\centering Average} & Vanilla & 25.40 & 37.44 & 20.93 & 20.46 & 3.09 & & {33.49} & {49.77} & {29.96} & \textbf{28.30} & 5.67 \\
                                & & & {CodeNER}& \textbf{35.94} & \textbf{48.32} & \textbf{41.01} & \textbf{23.07} & \textbf{12.04} & & \textbf{38.61} & \textbf{50.82} & \textbf{49.85} & {21.63} & \textbf{13.75} \\
\midrule
\multirow{15}{*}{\centering \textbf{Three-shot}} & \multirow{15}{*}{\centering Phi-3} & \multirow{2}{*}{\centering CoNLL03} & Vanilla & {59.36} & {78.55} & {68.29} & {55.70} & {2.58} & \multirow{15}{*}{\centering \shortstack{Llama-3 \\ 8B}} & {60.31} & {86.87} & {70.46} & {45.40} & {10.30}\\
                                & & & {CodeNER}& {52.26} & {73.12} & {55.99} & {40.61} & {22.92} & & {47.70} & {57.06} & {53.42} & 43.34 & 22.85\\
\cmidrule{3-9}\cmidrule{11-15}
                                & & \multirow{2}{*}{\centering FIN} & Vanilla & {29.24} & {35.53} & {24.24} & {11.76} & {0.40} & & {30.44} & {36.30} & {24.56} & {15.47} & {0.00}\\
                                & & & {CodeNER}& {39.08} & {53.44} & {33.85} & {14.29} & {1.65} & & {47.90} & {59.38} & {49.35} & 7.74 & 3.08 \\
\cmidrule{3-9}\cmidrule{11-15}
                                & & \multirow{2}{*}{\centering Arabic A} & Vanilla & {6.23} & {13.95} & {5.00} & {3.20} & {3.16} & & {19.27} & {49.76} & {5.21} & {13.19} & {12.54}\\
                                & & & {CodeNER} & {23.03} & {30.25} & {30.60} & {2.13} & {17.00} & & {29.84} & {44.79} & {37.59} & 0.00 & 20.69\\
\cmidrule{3-9}\cmidrule{11-15}
                                &  & \multirow{2}{*}{\centering Arabic B} & Vanilla & {4.98} & {8.21} & {4.97} & {5.00} & {2.64} & & {10.17} & {12.79} & {20.95} & {10.45} & {1.76}\\
                                & & & {CodeNER}& {22.86} & {33.44} & {27.78} & {5.94} & {15.42} & & {31.51} & {41.58} & {34.68} & 4.35 & 27.39 \\
\cmidrule{3-9}\cmidrule{11-15}
                                & & \multirow{2}{*}{\centering Finnish A} & Vanilla & {26.10} & {38.22} & {14.33} & {38.14} & {1.16} & & {37.90} & {51.64} & {41.40} & {43.91} & {8.06}\\
                                & & & {CodeNER}& {43.83} & {54.09} & {51.32} & {50.32} & 10.96 & & {40.17} & {39.02} & {57.77} & 42.51 & 14.63 \\
\cmidrule{3-9}\cmidrule{11-15}
                                & & \multirow{2}{*}{\centering DaNE} & Vanilla & {35.77} & {56.30} & {41.55} & {33.45} & {3.72} & & {48.73} & {71.98} & {46.88} & {52.07} & {11.18} \\
                                & & & {CodeNER}& {45.62} & {69.66} & {62.14} & {33.60} & 12.77 & & {42.53} & {49.90} & {60.83} & 43.10 & 16.29 \\
\cmidrule{3-9}\cmidrule{11-15}
                                & & \multirow{2}{*}{\centering Average} & Vanilla & {26.95} & {38.46} & {26.40} & \textbf{24.54} & {2.28} & & {34.47} & \textbf{51.56} & {34.91} & \textbf{30.08} & {7.31}\\
                                & & & {CodeNER}& \textbf{37.78} & \textbf{52.33} & \textbf{43.61} & {24.48} & \textbf{13.45} & & \textbf{39.94} & {48.62} & \textbf{48.94} & 23.51 & \textbf{17.49} \\
\bottomrule
\end{tabular}
}
% \end{adjustbox}
\caption{Experimental results of \textbf{one-shot} and \textbf{three-shot} for each label on \texttt{FIN}, \texttt{CoNLL03}, \texttt{Arabic A B}, \texttt{Finnish A}, and \texttt{DaNE} using the Phi- and Llama-3 models.}
\label{tab:few_shot_settings}
\end{table*}

\begin{table*}[t!]
\renewcommand{\arraystretch}{1}
% \begin{adjustbox}{width=1.80\columnwidth,center}
\centering
\Huge
\resizebox{0.9\textwidth}{!}{
\begin{tabular}{cccccccccccccc}
\toprule
\rowcolor{gray!10}
\textbf{Model} & \textbf{Dataset} & \textbf{Template} & {\textbf{$F_1$}} & {\textbf{PER}} & {\textbf{LOC}} &{\textbf{ORG}} & {\textbf{MISC}}  & \textbf{Model} & {\textbf{$F_1$}} & {\textbf{PER}} & {\textbf{LOC}} &{\textbf{ORG}} & {\textbf{MISC}}\\
\midrule
\multirow{41.5}{*}{\centering Phi-3} & \multirow{5}{*}{\centering FIN} & Vanilla & 1.24 & 1.82 & 0.00 & 0.00 & 0.00 & \multirow{41.5}{*}{\centering \shortstack{Llama-3 \\ 8B}} & 18.14 & 16.47 & 35.44 & {15.00} & 0.00 \\
                                & & {Vanilla w/ label (GPT)}& {1.85} & {2.71} & {0.00} & {0.00} & {0.00} & & {18.31} & {16.80} & {32.94} & \textbf{16.43} & {0.00} \\
                                & & {Vanilla w/ label (codeNER)}& {1.24} & {1.82} & {0.00} & {0.00} & {0.00} & & {18.40} & {16.47} & {36.36} & {15.64} & \textbf{1.75} \\
                                & & {CodeNER}& \textbf{17.51} & \textbf{18.18} & \textbf{27.12} & \textbf{10.43} & 0.00 & & {31.96} & {39.22} & {34.04} & 6.54 & 0.00 \\
                                & & {CodeNER w/o label}& {8.49} & {7.35} & {19.05} & {6.62} & 0.00 & & \textbf{45.62} & \textbf{57.26} & \textbf{41.12} & {9.40} & {1.33} \\
\cmidrule{2-8}\cmidrule{10-14}
                                & \multirow{5}{*}{\centering CoNLL03} & Vanilla & 11.73 & 21.20 & 15.44 & 3.75 & 0.00 & & {64.02} & {88.90} & {71.01} & \textbf{54.85} & 11.85 \\
                                & & {Vanilla w/ label (GPT)}& {11.92} & {21.50} & {15.33} & {3.86} & {0.83} & & \textbf{64.73} & \textbf{89.32} & \textbf{72.32} & {52.95} & \textbf{17.91} \\
                                & & {Vanilla w/ label (codeNER)}& {9.38} & {15.49} & {9.12} & {7.55} & {0.24} & & {64.24} & {89.13} & {71.25} & {53.89} & {14.76} \\
                                & & {CodeNER}& {36.57} & {54.03} & {44.64} & {19.75} & {16.95} & & 40.29 & 51.07 & 50.70 & 31.01 & {12.71} \\
                                & & {CodeNER w/o label}& \textbf{48.33} & \textbf{66.92} & \textbf{55.66} & \textbf{32.26} & \textbf{26.12} & & {37.00} & {44.27} & {46.17} & {31.80} & {10.75} \\
\cmidrule{2-8}\cmidrule{10-14}
                                & \multirow{5}{*}{\centering SwissNER} & Vanilla & 4.14 & 5.45 & 3.23 & 5.71 & 1.72 & & {49.04}& {72.91} & 58.82 & {52.06} & 6.15 \\
                                & & {Vanilla w/ label (GPT)}& {1.98} & {3.74} & {1.09} & {3.02} & {0.00} & & {52.28} & \textbf{73.06} & {67.23} & {52.91} & {8.00} \\
                                & & {Vanilla w/ label (codeNER)}& {1.63} & {7.41} & {0.00} & {1.02} & {0.00} & & \textbf{56.54} & {70.83} & \textbf{70.83} & \textbf{61.58} & \textbf{11.76} \\
                                & & {CodeNER} & {28.76} & {49.47} & \textbf{42.34} & {14.35} & {12.44} & & 35.48 & 28.99 & {62.37} & 30.55 & {6.55} \\
                                & & {CodeNER w/o label}& \textbf{31.78} & \textbf{50.55} & {41.67} & \textbf{21.71} & \textbf{15.76} & & {31.87} & {25.37} & {54.41} & {28.09} & {8.02} \\
\cmidrule{2-8}\cmidrule{10-14}
                                & \multirow{5}{*}{\centering Arabic A} & Vanilla & 0.00 & 0.00 & 0.00 & 0.00 & 0.00 & & {8.88} & 6.13 & 15.53 & {6.25} & 6.31 \\
                                & & {Vanilla w/ label (GPT)}& {0.00} & {0.00} & {0.00} & {0.00} & {0.00} & & {10.53} & {18.18} & {14.08} & \textbf{13.21} & {3.38} \\
                                & & {Vanilla w/ label (codeNER)}& {0.00} & {0.00} & {0.00} & {0.00} & {0.00} & & {16.80} & {25.40} & {20.10} & {7.55} & {11.11} \\
                                & & {CodeNER} & {5.04} & {10.71} & {4.19} & 0.00 & {3.21}  & & \textbf{27.52} & \textbf{36.30}  & \textbf{34.08} & 1.59 & \textbf{22.35} \\
                                & & {CodeNER w/o label}& \textbf{8.81} & \textbf{13.47} & \textbf{11.76} & \textbf{2.20} & \textbf{5.21} & & {21.84} & {31.29} & {21.05} & {9.14} & {19.05} \\
\cmidrule{2-8}\cmidrule{10-14}
                                & \multirow{5}{*}{\centering Arabic B} & Vanilla & 0.00 & 0.00 & 0.00 & 0.00 & 0.00 & & 4.62 & 5.78 & 5.33 & {9.68} & 2.41 \\
                                & & {Vanilla w/ label (GPT)}& {0.00} & {0.00} & {0.00} & {0.00} & {0.00} & & {9.87} & {11.70} & {14.01} & \textbf{17.72} & {4.60} \\
                                & & {Vanilla w/ label (codeNER)}& {0.00} & {0.00} & {0.00} & {0.00} & {0.00} & & {10.92} & {20.51} & {6.76} & {6.98} & {7.22} \\
                                & & {CodeNER} & {6.79} & {11.76} & {2.88} & \textbf{11.43} & \textbf{4.69} & & \textbf{27.52} & {36.30} & \textbf{34.08} & 1.59 & \textbf{22.35} \\
                                & & {CodeNER w/o label}& \textbf{9.09} & \textbf{13.53} & \textbf{17.07} & {0.00} &  {2.92} & & {24.69} & \textbf{36.31} & {31.15} & {5.85} & {16.18} \\
\cmidrule{2-8}\cmidrule{10-14}
                                & \multirow{5}{*}{\centering Finnish A} & Vanilla & 1.92 & 1.08 & 0.00 & 3.95 & 0.00 & & 23.80 & {42.90} & 20.74 & 27.99 & 1.71 \\
                                & & {Vanilla w/ label (GPT)}& {15.95} & {7.80} & {8.52} & {29.01} & {0.00} & & {25.91} & {46.21} & {13.77} & {34.50} & {3.00} \\
                                & & {Vanilla w/ label (codeNER)}& {13.35} & {7.88} & {2.79} & {25.97} & {0.00} & & {27.15} & \textbf{48.56} & {18.40} & {33.94} & {3.65} \\
                                & & {CodeNER} & {27.34} & {39.63} & {34.71} & {27.44} & \textbf{8.28} & & {29.46} & 31.27 & {42.03} & {29.02} & \textbf{14.11} \\
                                & & {CodeNER w/o label}& \textbf{31.69} & \textbf{42.40} & \textbf{44.13} & \textbf{32.61} & {5.95} & & \textbf{32.11} & {21.01} & \textbf{45.98} & \textbf{39.24} & {7.58} \\
\cmidrule{2-8}\cmidrule{10-14}
                                & \multirow{5}{*}{\centering DaNE} & Vanilla & 0.00 & 0.00 & 0.00 & 0.00 & 0.00 & & 13.49 & 17.35 & 9.43 & 18.09 & 4.84 \\
                                & & {Vanilla w/ label (GPT)}& {0.00} & {0.00} & {0.00} & {0.00} & {0.00} & & {21.80} & {22.79} & {30.22} & {29.86} & {2.92} \\
                                & & {Vanilla w/ label (codeNER)}& {0.00} & {0.00} & {0.00} & {0.00} & {0.00} & & {8.44} & {9.80} & {14.95} & {4.71} & {6.20} \\
                                & & {CodeNER} & {33.79} & {54.15} & {53.59} & {19.72} & {6.52} & & \textbf{37.55} & \textbf{42.29} & \textbf{59.05} & \textbf{36.13} & \textbf{15.33} \\
                                & & {CodeNER w/o label}& \textbf{41.27} & \textbf{61.10} & \textbf{58.33} & \textbf{30.48} & \textbf{12.57} & & {30.02} & {32.61} & {49.56} & {34.11} & {5.22} \\
\cmidrule{2-8}\cmidrule{10-14}
                                & \multirow{5}{*}{\centering Average} & Vanilla & 2.72 & 4.22 & 2.67 & 1.92 & 0.25 & & 26.00 & 35.78 & 30.90 & {26.27} & 4.75 \\
                                & & {Vanilla w/ label (GPT)}& {4.53} & {5.11} & {3.56} & {5.13} & {0.12} & & {29.06} & {39.72} & {34.94} & \textbf{31.08} & {5.69} \\
                                & & {Vanilla w/ label (codeNER)}& {3.66} & {4.66} & {1.70} & {4.93} & {0.03} & & {28.93} & \textbf{40.10} & {34.09} & {26.33} & {8.06} \\
                                & & {CodeNER} & {22.26} & {33.99} & {29.92} & {14.73} & {7.44} & & {31.81} & {37.74} & \textbf{43.38} & 19.44 & 
                                \textbf{12.23} \\
                                & & {CodeNER w/o label}& \textbf{25.64} & \textbf{36.47} & \textbf{35.38} & \textbf{17.98} & \textbf{9.79} & & \textbf{31.88} & {35.45} & {41.35} & {22.52} & {9.73} \\
\bottomrule
\end{tabular}}
% \end{adjustbox}
\caption{Experimental results for each label on \texttt{FIN}, \texttt{CoNLL03}, and \texttt{SwissNER} using the Phi- and Llama-3 models. Vanilla w/ label (GPT) and Vanilla w/ label (codeNER) each indicate that we added the corresponding labels explanation to the Vanilla prompt, labels explanation generated by GPT for the former, and labels explanation provided by CodeNER for the latter.}
\label{tab:result_label_vanilla}
\end{table*}

\begin{table*}[t!]
% \begin{adjustbox}{width=1.85\columnwidth,center}
\renewcommand{\arraystretch}{0.6}
\centering
\resizebox{0.6\textwidth}{!}{
\begin{tabular}{cccccc}
\toprule
\rowcolor{gray!10}
\multirow{-2}{*}{\textbf{Dataset}} & \multirow{-2}{*}{\textbf{Template}} & \multirow{-2}{*}{\textbf{Type}} & \multirow{-2}{*}{\textbf{Phi-3}} & \shortstack[c]{\textbf{Llama-3}\\\textbf{-8B}} & \shortstack[c]{\textbf{Llama-3}\\\textbf{-70B}} \\
\midrule
 \multirow{5}{*}{FIN} & \multirow{2}{*}{GoLLIE} & English & 8.26  & 9.21 & 11.02 \\
                                &  & Translation& 8.26 & 9.21 & 11.02 \\
\noalign{\vskip 1pt}
\cdashline{2-6}
\noalign{\vskip 3pt}
                                & \multirow{3}{*}{CodeNER}& {English}& \textbf{17.51}  & 31.96 & 19.39 \\
                                & & {Translation}& \textbf{17.51} & 31.96 & 19.39 \\
                                & & {w/o label}& 8.49 & \textbf{45.62} & \textbf{26.11}  \\
\cmidrule{1-6}
\multirow{5}{*}{CoNLL03} & \multirow{2}{*}{GoLLIE} & English & 35.39 & 34.72 & 53.93 \\
                                &  & Translation& 35.39 & 34.72 & 53.93 \\
\noalign{\vskip 1pt}
\cdashline{2-6}
\noalign{\vskip 3pt}
                                & \multirow{3}{*}{CodeNER}& {English}& 36.57 & \textbf{40.29} & \textbf{54.94} \\
                                & & {Translation}& 36.57 & \textbf{40.29} & \textbf{54.94} \\
                                & & {w/o label}& \textbf{48.33} & 37.00 & 47.53 \\
\cmidrule{1-6}
\multirow{5}{*}{SwissNER} & \multirow{2}{*}{GoLLIE} & English & \textbf{32.54} & 24.98 & 44.49 \\
                                &  & Translation& 27.59 & 10.04 & 39.45 \\
\noalign{\vskip 1pt}
\cdashline{2-6}
\noalign{\vskip 3pt}
                                & \multirow{3}{*}{CodeNER}& {English}& 32.16 & 35.03 & \textbf{49.12} \\
                                & & {Translation}& 28.76 & \textbf{35.48} & 47.04 \\
                                & & {w/o label}& 31.78 & 31.87 & 39.21 \\
\cmidrule{1-6}
\multirow{5}{*}{Arabic A} & \multirow{2}{*}{GoLLIE} & English & \textbf{14.60} & 12.57 & 22.83 \\
                                &  & Translation& 3.07 & 7.57 & 21.72 \\
\noalign{\vskip 1pt}
\cdashline{2-6}
\noalign{\vskip 3pt}
                                & \multirow{3}{*}{CodeNER}& {English}& 9.18 & 16.24 & \textbf{29.47}	\\
                                & & {Translation}& 5.04 & 20.42 & 28.83 \\
                                & & {w/o label}& 8.81 & \textbf{21.84} & 29.00 \\
\cmidrule{1-6}
\multirow{5}{*}{Arabic B} & \multirow{2}{*}{GoLLIE} & English & \textbf{15.83} & 16.53 & 25.12 \\
                                &  & Translation& 2.35 & 8.70 & 25.42 \\
\noalign{\vskip 1pt}
\cdashline{2-6}
\noalign{\vskip 3pt}
                                & \multirow{3}{*}{CodeNER}& {English}& 9.31 & 16.58 & \textbf{33.78} \\
                                & & {Translation}& 6.79 & \textbf{27.52} & 32.16 \\
                                & & {w/o label}& 9.09 & 24.69 & 33.62 \\
\cmidrule{1-6}
\multirow{5}{*}{Finnish A} & \multirow{2}{*}{GoLLIE} & English & 23.48 & 21.86 & 34.15 \\
                                &  & Translation& 19.61 & 6.58 & 26.72 \\
\noalign{\vskip 1pt}
\cdashline{2-6}
\noalign{\vskip 3pt}
                                & \multirow{3}{*}{CodeNER}& {English}& \textbf{32.37} & \textbf{33.29} & \textbf{53.16} \\
                                & & {Translation}& 27.34 & 29.46 & 48.89 \\
                                & & {w/o label}& 31.69 & 32.11 & 48.98 \\
\cmidrule{1-6}
\multirow{5}{*}{DaNE} & \multirow{2}{*}{GoLLIE} & English & 27.37 & 25.33 & \textbf{47.17} \\
                                &  & Translation& 21.82 & 17.11 & 43.68 \\
\noalign{\vskip 1pt}
\cdashline{2-6}
\noalign{\vskip 3pt}
                                & \multirow{3}{*}{CodeNER}& {English}& 38.25 & 36.53 & 42.50 \\
                                & & {Translation}& 33.79 & \textbf{37.55} & {42.93} \\
                                & & {w/o label}& \textbf{41.27} & 30.02 & 35.06 \\
\cmidrule{1-6}
\multirow{5}{*}{Average} & \multirow{2}{*}{GoLLIE} & English & 22.50 & 20.74 & 34.10 \\
                                &  & Translation& 16.87 & 13.42 & 31.71 \\
\noalign{\vskip 1pt}
\cdashline{2-6}
\noalign{\vskip 3pt}
                                & \multirow{3}{*}{CodeNER}& {English}& \textbf{25.05} & 29.99 & \textbf{40.34} \\
                                & & {Translation}& 22.26 & 31.81 & 39.17 \\
                                & & {w/o label}& 25.64 & \textbf{31.88} & 37.07 \\
\bottomrule
\end{tabular}}
% \end{adjustbox}
\caption{Experimental macro F1-scores for \texttt{GoLLIE} and \texttt{CodeNER} using label descriptions either translated into each dataset's language or kept in English. In the type column, \texttt{Translation} denotes labels translated with Google Translator, while \texttt{English} indicates the original prompt remained unchanged.}
\label{tab:translate_results}
\end{table*}

\end{document}